\documentclass[sigconf]{acmart}

\AtBeginDocument{%
  }

\setcopyright{acmlicensed}

\acmYear{2025}\copyrightyear{2025}
\acmConference[COMPASS '25]{ACM SIGCAS/SIGCHI Conference on Computing and Sustainable Societies}{July 22--25, 2025}{Toronto, ON, Canada}
\acmBooktitle{ACM SIGCAS/SIGCHI Conference on Computing and Sustainable Societies (COMPASS '25), July 22--25, 2025, Toronto, ON, Canada}
\acmDOI{10.1145/3715335.3735488}
\acmISBN{979-8-4007-1484-9/25/07}

\usepackage[textsize=tiny]{todonotes}

\newcommand{\NoteMatt}[2][]{}

\newcommand{\NoteJohn}[2][]{}

\newcommand{\NoteMihir}[2][]{}

\newcommand{\NoteAsmita}[2][]{}

\newcommand{\NoteMonica}[2][]{}

\newcommand{\NoteJunia}[2][]{}

\usepackage{array}
\usepackage{url}
\usepackage{multirow}
\usepackage{graphicx}
\usepackage{subcaption}
\usepackage{float}

\begin{document}

\title{Predicting the Past: Estimating Historical Appraisals with OCR and Machine Learning}

\author{Mihir Bhaskar}
\email{mihirb@stanford.edu}
\orcid{https://orcid.org/0009-0006-9104-6988}
\authornote{Affiliation is for identification purposes only. Work was done in an individual capacity.}
\affiliation{%
  \institution{Stanford University}
  \city{Stanford}
  \state{CA}
  \country{USA}
}

\author{Jun Tao Luo}
\email{jtluo@alumni.cmu.edu}
\orcid{https://orcid.org/0000-0002-2681-8922}
\affiliation{%
  \institution{Carnegie Mellon University}
  \city{Pittsburgh}
  \state{PA}
  \country{USA}
}

\author{Zihan Geng}
\email{zihangen@alumni.cmu.edu}
\orcid{https://orcid.org/0009-0008-2325-7366}
\affiliation{%
  \institution{Carnegie Mellon University}
  \city{Pittsburgh}
  \state{PA}
  \country{USA}
}

\author{Asmita Hajra}
\email{ahajra@alumni.cmu.edu}
\orcid{https://orcid.org/0009-0001-7700-7492}  %
\affiliation{%
  \institution{Carnegie Mellon University}
  \city{Pittsburgh}
  \state{PA}
  \country{USA}
}

\author{Junia Howell}
\email{jhowel4@uic.edu}
\orcid{https://orcid.org/0000-0003-4631-7225}
\affiliation{%
  \institution{University of Illinois Chicago}
  \city{Chicago}
  \state{IL}
  \country{USA}
}

\author{Matthew R. Gormley}
\email{mgormley@cs.cmu.edu}
\orcid{https://orcid.org/0000-0003-4207-5986}
\affiliation{%
  \institution{Carnegie Mellon University}
  \city{Pittsburgh}
  \state{PA}
  \country{USA}
}

\renewcommand{\shortauthors}{Anonymous}

\begin{abstract}

Despite well-documented consequences of the U.S. government's 1930s housing policies on racial wealth disparities, scholars have struggled to quantify its precise financial effects due to the inaccessibility of historical property appraisal records. Many counties still store these records in physical formats, making large-scale quantitative analysis difficult.
We present an approach scholars can use to digitize historical housing assessment data, applying it to build and release a dataset for one county. Starting from publicly available scanned documents, we manually annotated property cards for over 12,000 properties to train and validate our methods. We use OCR to label data for an additional 50,000 properties, based on our two-stage approach combining classical computer vision techniques with deep learning-based OCR.
For cases where OCR cannot be applied, such as when scanned documents are not available, we show how a regression model based on building feature data can estimate the historical values, and test the generalizability of this model to other counties. With these cost-effective tools, scholars, community activists, and policy makers can better analyze and understand the historical impacts of redlining.

\end{abstract}



\maketitle

\newcommand{\ourFigureWidth}{\columnwidth}

\section{Introduction} \label{introduction}

Historians and social scientists have repeatedly shown that the U.S. government's 1930s housing policies exacerbated racial residential segregation \cite{Jackson1985,Rothstein2017,Faber2020}. Colloquially called redlining, these policies introduced appraisal practices that used a neighborhood's racial composition to determine a property's value \cite{HowellKorverGlenn2021,WinlingandMichney2021,Michney2022}. However, scholars have yet to enumerate the impact of these changes on racial wealth gaps or outline potential approaches to remedy these inequities. A key barrier to this work is the inaccessibility of historical records.

\begin{figure}[b]
    \centering
    \includegraphics[width=\columnwidth]{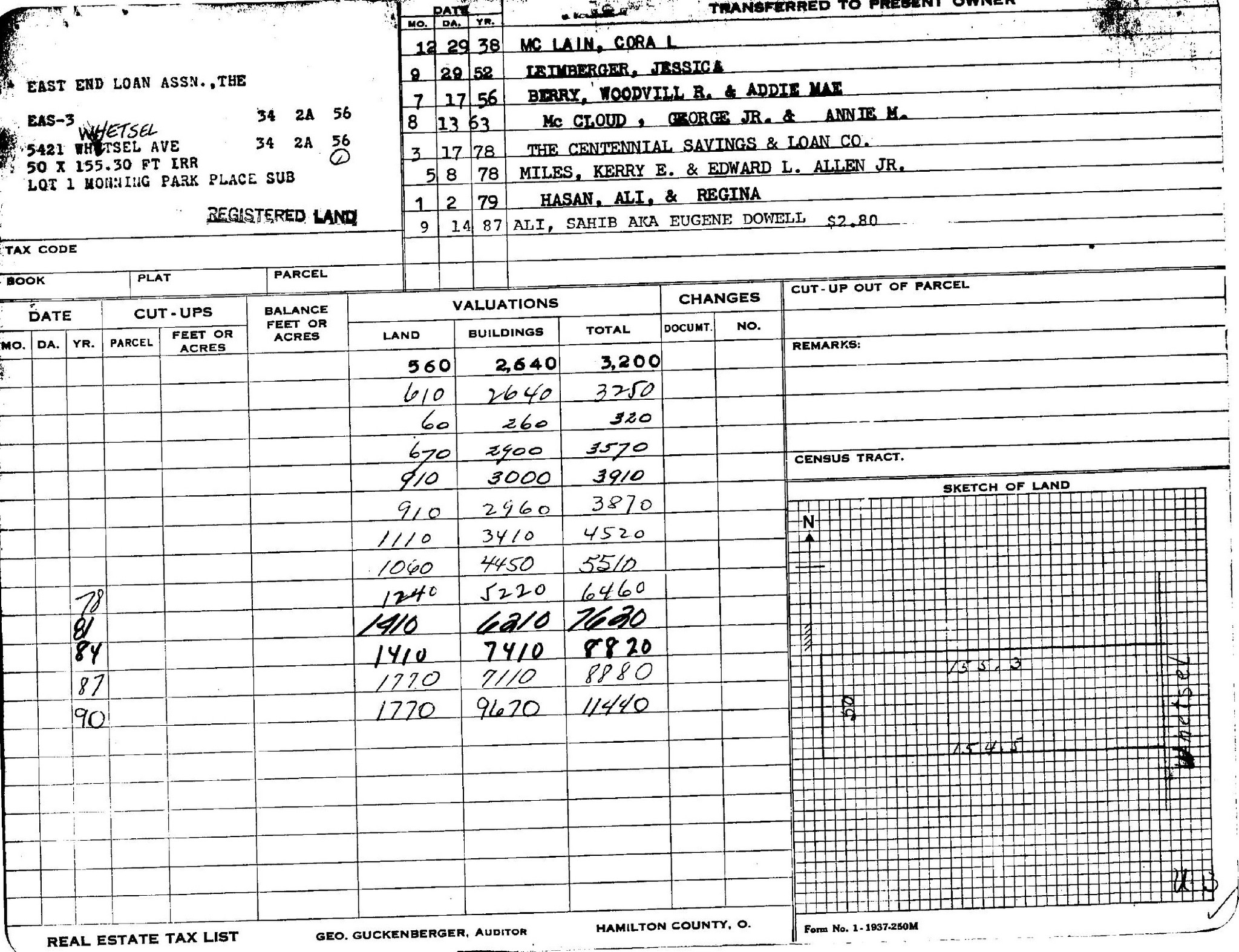}
    \centering
    \caption{Example property assessment card}
    \label{fig:sample-ownership}
\end{figure}

Most U.S. counties have detailed and comprehensive historical property data kept by the government assessor, recorder, and planning offices. Yet many of these files are still stored in physical formats with handwritten information (e.g. Figure \ref{fig:sample-ownership}), curtailing quantitative analyses. In this paper, we present a cost-effective and time-efficient approach for deriving county-wide historical estimates of property values at the building level. We use this approach to create a novel dataset of digitized historical housing assessment data for one county (Hamilton County, Ohio), and explore its generalizability to others.

Our approach entails two steps: (1) extracting historical values from scanned documents using OCR, and (2) extrapolating values for all properties based on a regression model. With this relatively cheap and quick method, scholars, community activists, and elected officials can empirically demonstrate the extent to which these policies influenced local communities and what could be done to mitigate their ongoing detrimental impacts.

Our code\footnote{\url{https://github.com/JunTaoLuo/ErukaExp}} and dataset\footnote{\url{https://huggingface.co/datasets/eruka-cmu-housing/historical-appraisals-ocr-ml}} are publicly available.

\subsection{The Racialization of Appraising Methodology} \label{historical_background}

At the beginning of the 20th century, state and local governments relied on real estate assessments to collect property taxes--their primary source of revenue \cite{Carlson2005}. Unlike contemporary property assessments that use a \emph{sales comparison} approach, relying on market values of similar properties sold recently, these historic values were based on the estimated \emph{construction cost} of the dwelling. To systemize property values, government assessors increasingly relied on published books that provided tables to help calculate the cost of homes based on their features (e.g., two bedroom, one bathroom, brick exterior). Although property assessments did not always match the price a house might sell for on the market, prices were often comparable as many market appraisers also used a similar cost approach \cite{Michney2022}. That is, until the federal government transformed market appraising practices.

A central component of President Franklin D. Roosevelt's plan to address the economic and housing crisis caused by the 1929 Stock Market Crash was reshaping property mortgages. Up to this point, U.S. mortgages required 50-60 percent down payments, lasted one to five years, and only mandated the borrower pay interest payments until the loan duration was over and the remaining principle was due \cite{Stuart2003, Szto2013}. This financing structure contributed to the cascading bank failures. The Roosevelt Administration sought to stabilize the housing market by introducing and normalizing amortized, 15-year mortgages that only required 10 percent down payments \cite{Marchiel2020}. To get private lenders to adopt this radically new type of loan, the 1934 National Housing Act introduced federally-backed mortgage insurance that passed much of the risks of default onto the federal government  \cite{Jackson1985, Stuart2003, Marchiel2020}. There was just one catch. Qualifying for the federal mortgage insurance required obtaining a certified appraisal that complied with federal standards.

To ensure a consistent appraising approach, the newly formed Federal Housing Administration hired Frederick Babcock to write the first federal underwriting manual, which was published in 1936 \cite{WinlingandMichney2021}. Like many other White real estate scholars at the time, Babcock was an eugenicist who believed that White communities were the most valuable because White people were the most evolved "race" \cite{Marchiel2020}. Babcock infused the federal underwriting manual with these ideas. Instead of building off existing assessment precedent and using a cost-based approach, Babcock elevated the importance of the area's racial and socioeconomic composition above property features \cite{Stuart2003, Marchiel2020, WinlingandMichney2021}. This combined with federally created color-coded maps of ``risk levels'' by neighborhood began transforming market appraisal values across the country. By the 1940s, similar homes in White neighborhoods and communities of color went from being comparably priced to appraising at radically different values.

Social scientists have repeatedly documented the devastating consequences that this federal policy had on racial inequality in wealth, health, and housing \cite{Jackson1985,Rothstein2017,Faber2020,Marchiel2020,HowellKorverGlenn2021}. Yet, scholars have been unable to quantify the monetary impact on individual families and communities because the necessary historical parcel-level assessment data has not been digitized.

\subsection{Technical Data Challenges}

There are two primary challenges faced by scholars interested in using this historical assessment data: (1) accurately digitizing tabulated, handwritten information and (2) obtaining and scanning the physical cards.

Many historical assessment records were kept on physical index cards with tabulated hand written values. The tabular structure of these documents creates a technical challenge for digitization. Recent advancements in historical document digitization have innovated new ways to extract data from scanned documents including balance sheets \cite{CORREIA2023101475}, newspapers \cite{booth-etal-2022-language, 10.1007/978-3-030-19823-7_30}, historical censuses \cite{muellermulti}, and church records \cite{9775079}. Yet, this work does not address the challenge of semantic understanding of tabular documents. Accurately capturing tabulated data requires identifying the text positions by using supporting context, similar to approaches in described in tabular OCR works \cite{DBLP:journals/corr/abs-2010-08591, fischer2021multi, Ranjan2021}. To do this for property assessment cards, we need a card-specific matching approach that could identify value locations prior to attempting character recognition.

The second challenge for scholars seeking to use these historical data is the physical accessibility of the records. Although most local governments are required by law to keep their historical property records, many of these records are still kept in physical filing cabinets. Scanning hundreds of thousands of index cards is time-consuming and cost prohibitive. We need a method that can estimate the historical values using other readily available data sources.

\subsection{Project scope and process}

Our work has two outcomes: first, a dataset of 1933 assessment values for one county (\S \ref{sec:new-dataset}) and, second, a regression model for predicting 1930s assessment values based on property features (\S \ref{sec:regression-models-for-housing}).

The primary objective of this project is to provide property value estimates for the years imminently preceding Frederick Babcock's Federal Underwriting Manual, published in 1936. We focus on Hamilton County, Ohio (primarily the city of Cincinnati), because scanned images of all historical property cards have been made publicly available. In the case of Hamilton County, obtaining the pre-1936 value simplifies the problem to retrieving a single-cell in the table---specifically, the earliest assessment on the card, which was conducted in 1933 in most cases. In addition to this, we present a method to parse and retrieve the entire historical property card in a ``comprehensive" format, providing additional context and data for researchers interested in the full historical record, including values after 1933. Section \S \ref{sec:new-dataset} lays out our process of compiling this dataset of assessment values, both single-cell and comprehensive, for Hamilton county. We hand-annotate over 70,000 table cells to produce highly reliable train and test data. We then use computer vision techniques to identify value placements and use optical character recognition (OCR) models to extract desired amounts.

\par

To address the problem of historical records that have not been scanned, we present a regression-based approach to digitization. As described in Section \S \ref{introduction}, the pre-1936 (i.e., pre-Babcock) property assessments were conducted based on building features in a relatively standardized way. This standardization offers hope for learning a generalized estimation of values based on building feature data, even in the absence of physical cards, to obtain pre-1936 values for other counties that may not have scanned records. Since we do not observe building features in the 1930s, we use contemporary data as a proxy, on the assumption that it is correlated with the historical values. Recent building and parcel information is typically available in digitized format for most counties through the county assessor. We combine this feature data with the hand annotations and OCR values from Hamilton County in 1933 to train and validate a regression model for pre-1936 values. We test the generalizability of this method by running the model on data from Franklin County (primarily Columbus), Ohio, and comparing our estimates with their publicly available historical records. This regression-based modeling approach is detailed in Section \S \ref{sec:regression-models-for-housing}.
\par
\begin{figure}[t]
    \centering
    \includegraphics[width=\columnwidth]{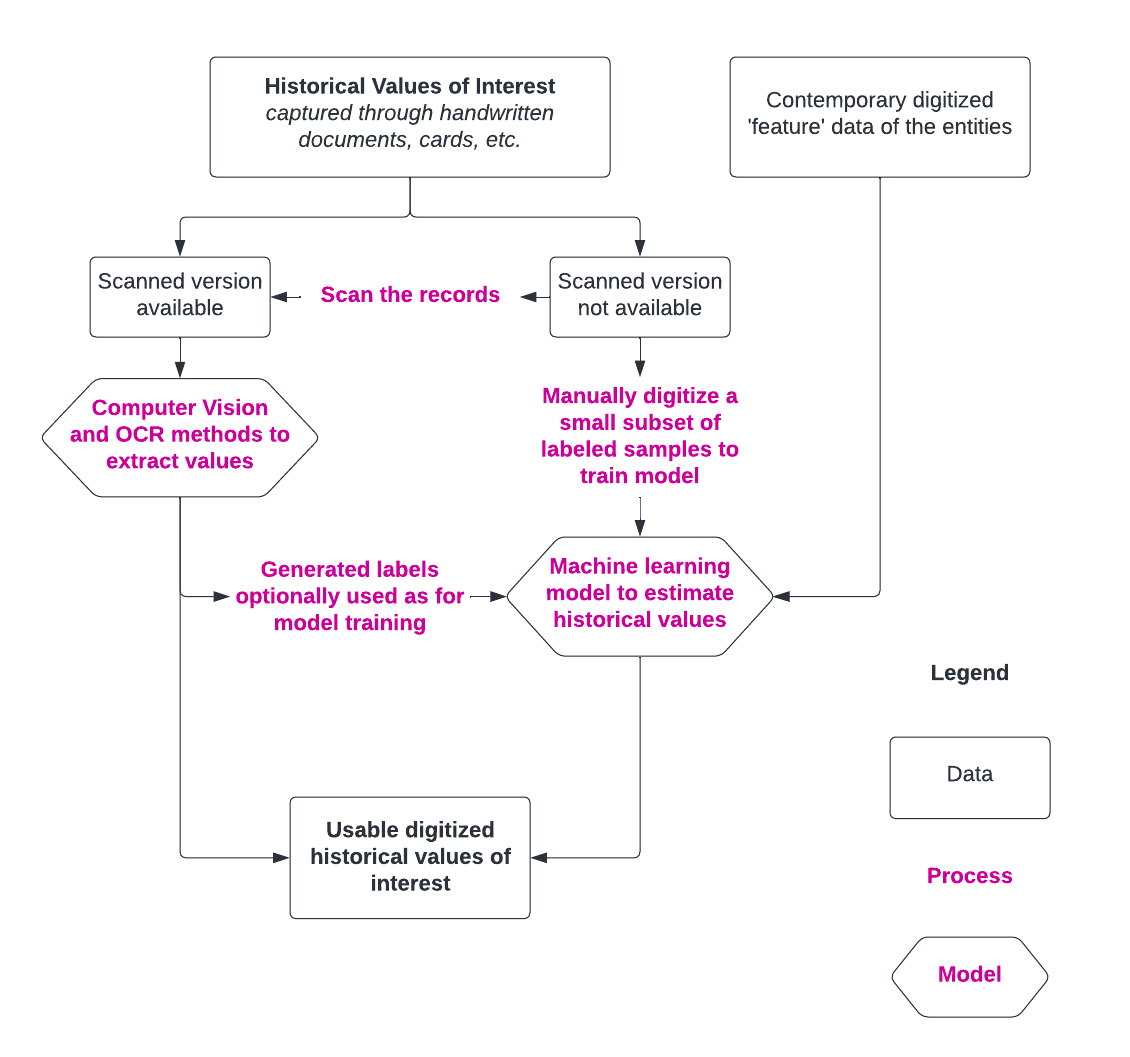}
    \centering
    \caption{Proposed methodology for digitizing historical property records}
    \label{fig:proposed-methods}
\end{figure}

Our overall approach is summarized in Figure \ref{fig:proposed-methods}.

\section{Building a Dataset of Historical Housing Assessments for Hamilton County}
\label{sec:new-dataset}

We introduce a novel dataset derived from scanned housing records in Hamilton County (Cincinnati), Ohio (\S \ref{sec:property-ownership-cards}), annotated in both comprehensive and single-cell formats (\S \ref{sec:manual-annotation}). Our multi-stage extraction approach segments property cards into table cells, applies OCR to identify values (\S \ref{sec:ocr-methodology}), and achieves greater accuracy than state-of-the-art document understanding methods despite its simplicity (\S \ref{sec:results-ocr}).

\subsection{Related Work}
\label{sec:dataset-related-work}

\subsubsection{Document Layout Understanding}
Recent research in document understanding has focused primarily on cases where the structure/formatting of each document may differ and is unknown \textit{a priori}. Unified networks, such as TRIE \citep{trie}, combine text detection with information extraction, enhancing the processing of complex documents like invoices and resumes. For document image understanding in this setting, pre-training a joint model of text and layout, such as in LayoutLM \cite{layoutlm} or LayoutLMv3 \cite{layoutlmv3}, can yield strong capabilities in generalizing to layouts not seen during training. In parallel, the LayoutParser framework \cite{shen2021layoutparserunifiedtoolkitdeep} has emerged as a standard modular tool for document segmentation and layout analysis, adopted across disciplines for its flexibility and ease of integration into OCR pipelines. It provides a unified interface for applying deep learning-based object detection models (e.g., Detectron2) to detect layout elements such as text blocks, tables, and figures directly from document images. While these unified methods can be applied in our setting (\ref{sec:results-ocr} discusses our attempts to do so), our custom segmentation/OCR approach capitalizes on a fixed, known layout that is identical across all property cards within a county.

\subsubsection{Image Alignment}
Since the layout of the Hamilton county property cards is known ahead of time, this enables us to align it to a fixed template and know, with high precision, the position of each table cell. Image alignment has rich history in computer vision, yielding robust methods that handle varying degrees of distortion, rotation, and scaling. Key techniques involve feature detection and matching algorithms, such as Scale-Invariant Feature Transform (SIFT) \cite{sift}, Speeded Up Robust Features (SURF) \cite{surf}, and ORB \cite{rublee_orb_2011} which have proven effective in identifying corresponding points between images. Further, the use of homography matrices for transforming the perspective of images has been instrumental in achieving precise alignment. Our work follows prior work on aligning to a fixed template before performing OCR \cite{mahajan_character_2019,pant_use_2022}.

\subsubsection{Optical Character Recognition (OCR)}

Approaches for optical character recognition (OCR) range widely from classical techniques to modern deep learning approaches. The TesseractOCR engine \cite{tesseractocr} uses classical methods including line finding, feature-based methods, and adaptive classifiers. Modern approaches include the TrOCR model \citep{trocr}, which performs text recognition by integrating pre-trained image and text Transformer models. This approach marks a departure from earlier deep learning methods that relied on CNNs for image understanding and RNNs for character-level text generation. Though the property card template is machine printed, most of the data we wish to extract from them is handwritten. Recognition of handwriting is particularly challenging for OCR models, and typically requires specialized training datasets \cite{memon_handwritten_2020}.

\subsection{Historical Property Assessment Cards}
\label{sec:property-ownership-cards}

We develop this dataset from historical property assessment cards from Hamilton County, Ohio. These cards were maintained by the elected auditors who were responsible for calculating property assessments. Since we are primarily interested in the 1933 assessment value of residential homes, we restrict our examination to residential parcels \footnote{Parcels are the formal word in real estate for the physical property boundary. We use parcel interchangeably with properties.} with a single building constructed prior to 1930.

The resulting set of 59,378 parcels forms the overall sample of interest for Hamilton County. Of this set, we were only able to successfully retrieve 56,037 scanned documents, which indicates 5.6\% of the parcels do not have publicly available scanned documents. To ensure we do not introduce bias due to these missing documents, we perform a classifier two-sample test as detailed in Appendix \ref{app:missing-oc}. Next, we perform basic pre-processing on the documents including cropping, rotating, and conversion to grayscale.

\subsection{Manual Annotation}
\label{sec:manual-annotation}

We used Upwork to hire contractors to manually label a subset of our samples at a rate of 12.50 US\$ - 15 US\$ an hour. We produced two types of annotation: a \textit{comprehensive} format and a \textit{single-cell} format.

\begin{figure}[t]
    \centering
    \includegraphics[width=\columnwidth]{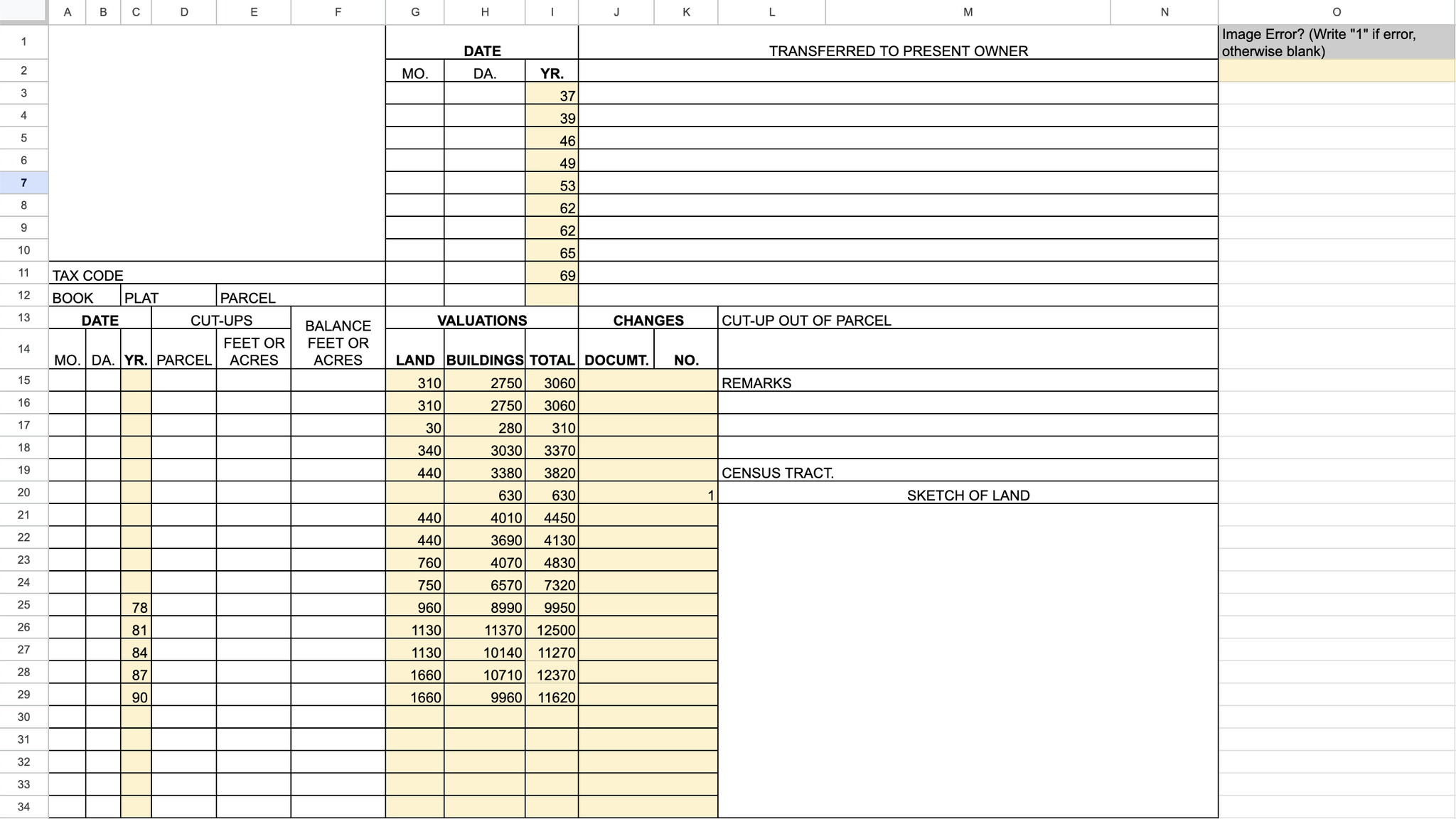}
    \centering
    \caption{Example Manual Annotation of Property Assessment Card}
    \label{fig:example_annotation}
\end{figure}

For the \textit{comprehensive} format, we sampled 588 property cards from Hamilton county, and the annotators labeled the dollar valuations for the land, building, and total columns as well as the corresponding year; they also annotated the year of each transfer (refer Figure \ref{fig:sample-ownership}). This annotation covered 61,816 table cells, of which 35,661 (57\%) were nonempty. This annotation was comprehensive with the exception that we did not annotate month/day of dates nor the owner name. An example annotation is Figure \ref{fig:example_annotation}.
Two of the authors and the three contractors all annotated the same small sample of property cards in order to assess inter-annotator agreement; the disagreement was below 0.02 for every pair of annotators.

\begin{figure}[t]
    \centering
    \includegraphics[width=\columnwidth]{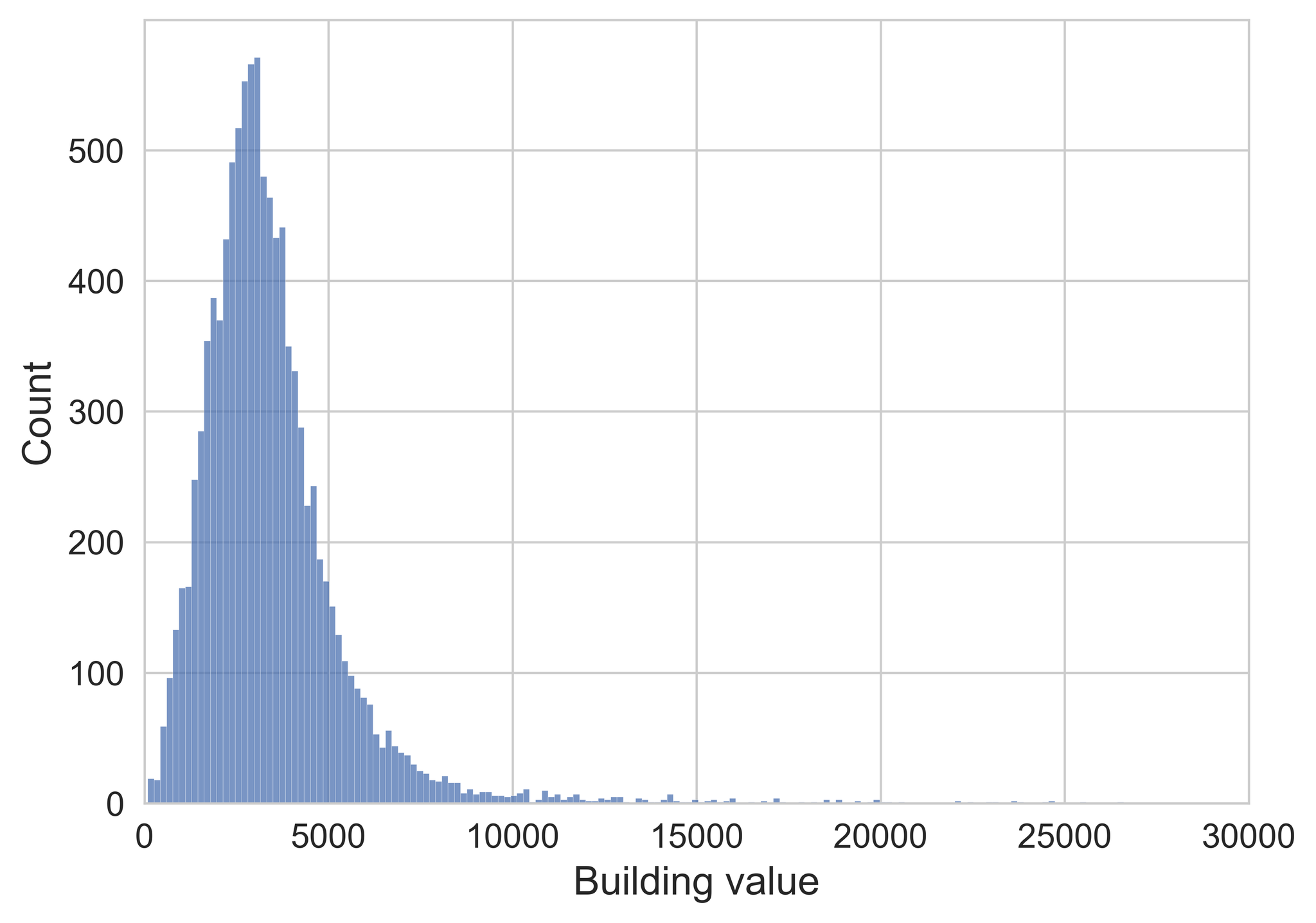}
    \centering
    \caption{Histogram of building valuations in US\$ circa 1933, Hamilton county}
    \label{fig:target hist}
\end{figure}

\begin{figure}[t]
    \centering
    \includegraphics[width=\columnwidth]{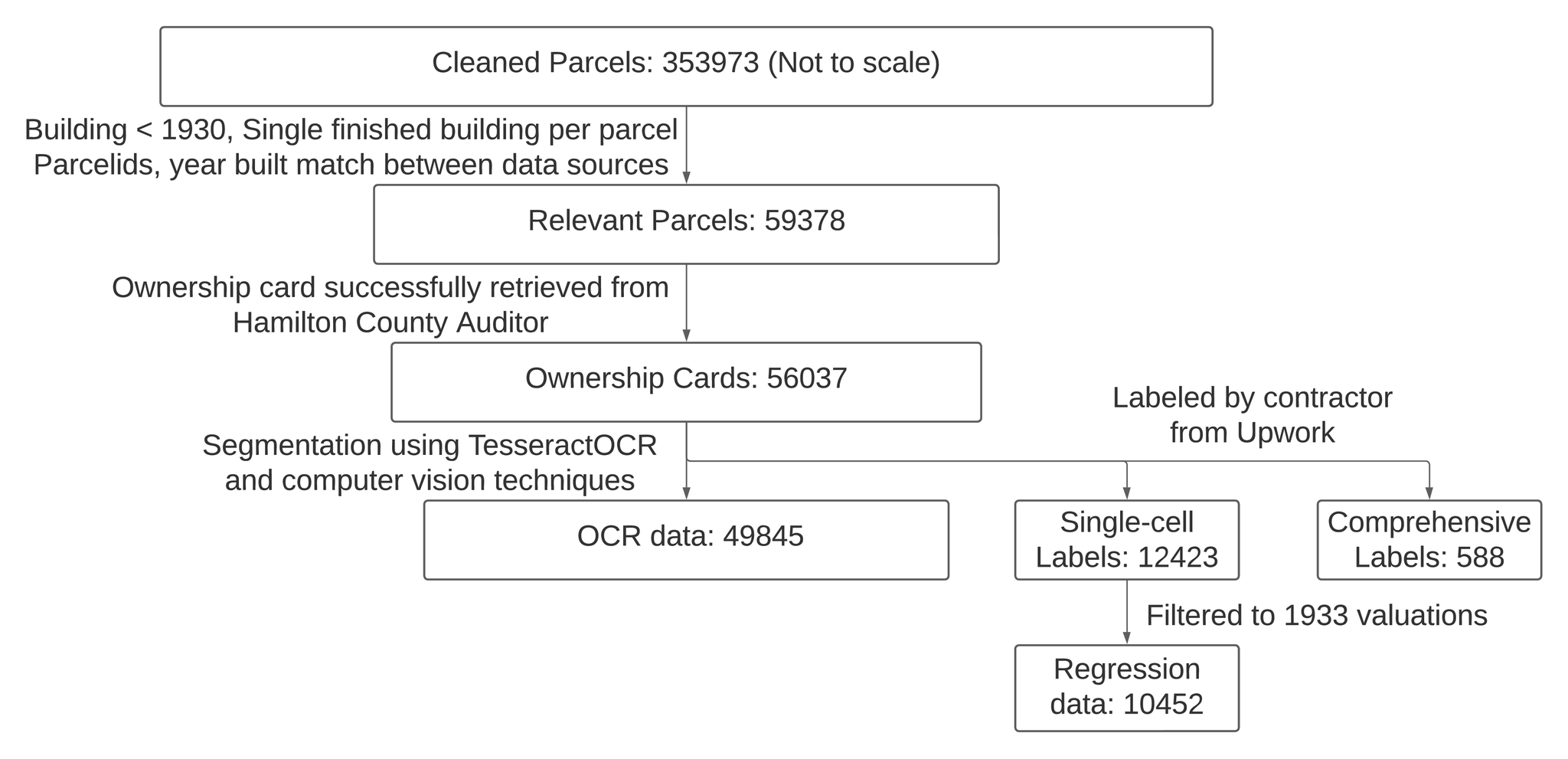}
    \centering
    \caption{Data processing flow for Hamilton county}
    \label{fig:data-funnel}
\end{figure}

For the \textit{single-cell} format, a much broader set of 12,423 properties were annotated for only a single building valuation, year, and whether the value was handwritten. This was a random sample of cards for which the first value in the "BUILDING" column was recorded. In most cases, where the ``year" column is blank, this refers to the 1933 assessment value. After filtering down to those with a 1933 valuation, 10,452 samples remained. The distribution of the annotated value, based on the 10,452 samples, is shown in Figure \ref{fig:target hist}.
Figure \ref{fig:data-funnel} depicts the data preparation process and the number of samples after each step.

\subsection{OCR Methodology}
\label{sec:ocr-methodology}
\label{cv ocr methodology}

Although there are many off-the-shelf solutions for document parsing and OCR, we did not find an existing method that was able to accurately extract values from the tables in our documents. We tried TesseractOCR, LayoutLMv3, LayoutParser, Microsoft Azure AI Document Intelligence, and OpenAI's ChatGPT models - challenges of which are noted in (\S \ref{sec:results-ocr}). As such, we developed a two-stage solution for identifying table cells (\S \ref{sec:methods-segmentation}) which were individually processed by OCR (\S \ref{sec:methods-ocrmodels}).

\subsubsection{Tabular Data Segmentation Methods}
\label{sec:methods-segmentation}

The first challenge is to recognize the tabular structure of the assessment card documents and locate the relevant information.

For the \textit{comprehensive} format, we align scanned property assessment cards to an empty reference template using a homography matrix derived from ORB-based feature matching \cite{rublee_orb_2011}. Both images are converted to grayscale to minimize lighting variations, and dark areas are brightened for better alignment. Feature matching is performed via OpenCV’s \cite{opencv_library} BFMatcher with Hamming distance, retaining the top 5\% of 5,000 matches after sorting by score. To enhance reliability, we filter matches by enforcing quadrant consistency. The homography matrix, computed with RANSAC \cite{fischler_random_1981}, aligns the scanned images, enabling precise table cell extraction of key fields (valuations, year).

For the \textit{single-cell} format, given that we use the building value of the first recorded appraisal in the Hamilton property records, this involves obtaining the first entry of the ``BUILDINGS" column. To accomplish this, we use a customized process for segmentation which involves using TesseractOCR to locate the column header ``BUILDINGS", and then using Hough Transform to locate surrounding row and column divisions for cropping individual table cells. For more details about the segmentation step, see Appendix \ref{app:segmentation-tesseract}.

The individual cropped cell's images (of both the comprehensive and single-cell variety) are then passed to the OCR models.

\subsubsection{Optical Character Recognition (OCR) Models}
\label{sec:methods-ocrmodels}

Given that our task involves recognizing a mix of handwritten and typed numerical values, it is challenging to use off-the-shelf OCR solutions or pre-trained models such as TesseractOCR or TrOCR since these models predict \emph{all} characters, including digits, punctuation and letters, and perform poorly on our noisy dataset. Initial experiments show that these pre-trained models would often confuse letters and digits, including recognizing the digit 0 with the letter O and the digit 1 with lowercase L or uppercase I. To address this type of error, we perform additional fine-tuning on our model of choice - TrOCR - using different datasets for the \textit{single-cell} format and the \textit{comprehensive} format.
For the \textit{comprehensive} format, we used our annotated dataset of selected columns of numerical values of the property cards.
For the \textit{single-cell} format, we used a mixture of different datasets including CAR-B (handwritten digit strings from scanned checks) \cite{6981115} and DIDA (historical handwritten digit dataset) \cite{Kusetogullari2020DIDA}. These supplemented our Hamilton County card annotations.
For additional details about our OCR experiments, see Appendix \ref{app:ocr}.

\subsection{OCR Results}
\label{sec:results-ocr}

\subsubsection{Experiments with existing tools}

We piloted both open source methods and commercial solutions (parenthetical results are from small scale experiments). TesseractOCR \cite{tesseractocr} struggled to accurately identify the table structure and did OCR poorly on the handwritten numbers -- detailed experiments are in Appendix \ref{app:segmentation-tesseract} and \ref{app:ocr-tesseract}. Using LayoutLMv3 \cite{layoutlmv3} for pre-processing offered improvements, but still led to poor identification of table cells ($\sim$67\% accuracy). LayoutParser \cite{shen2021layoutparserunifiedtoolkitdeep} resulted in incorrect identification of table cells (<20\% accuracy) as well, refer Appendix \ref{app:segmentation-layoutparser} for more details. Switching to a Microsoft Azure AI Document Intelligence showed marked improvement on OCR ($\sim$94\% accuracy), but it still struggled to correctly identify table cells ($\sim$78\% accuracy), and its price was prohibitively expensive (e.g. 1600 US\$ for $\sim$56,000 property cards from just one county).

To benchmark against an approach more likely to be adopted by social scientists with minimal OCR expertise, we also conducted small-scale experiments on ChatGPT - specifically the GPT-4o model \cite{openai2024gpt4o}. While GPT-4o offers an appealing low-effort alternative, we found it to be a hit-or-miss solution due to variance in its performance when applied to large-scale, one-shot property card understanding—i.e., image analysis on the entire card at once. While these models are very intelligent, achieving accuracy to the level of our specialized approach in a one-shot setting would require extensive prompt engineering by a data scientist or a social scientist, and post-processing of results to store in appropriate data formats - which would defeat the purpose of using this tool as a quick non-technical solution. However, when using the ChatGPT API in a more constrained setting — such as OCR on cropped subsections of the card after segmentation (e.g., individual cell value), the accuracy of OCR matches or exceeds that of the fine-tuned TrOCR model. The cost of using GPT-4o in this way can scale quickly for very large datasets (e.g. from 600 US\$ for $\sim$56,000 property cards from one county, to 60,000 US\$ for 100 counties, compared to the one-time investment of developing a custom OCR model). Moreover, the model has practical limitations: no private deployment options, latency, and rate limits that complicate batch processing. These trade-offs, while acceptable and well suited for exploratory use, make GPT-4o less suitable for very high-volume workflows. Performance and cost details are provided in Appendix \ref{app:segmentation-chatgpt} and Appendix \ref{app:costs}.

\subsubsection{Tabular Data Segmentation Results}
\label{sec:results-segmentation}

\begin{figure}
    \centering
    \includegraphics[width=\columnwidth]{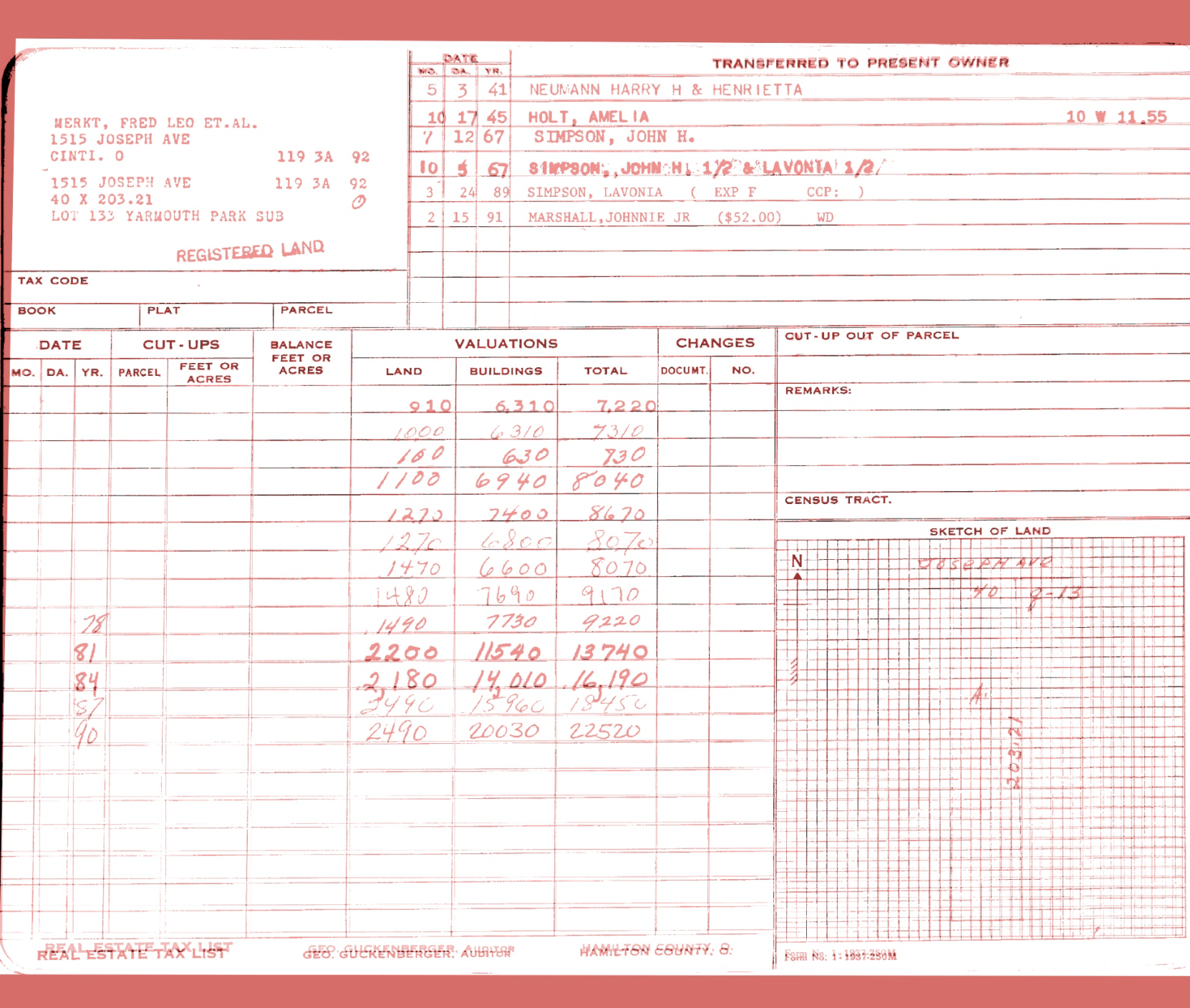}
    \caption{Aligned property card (red) overlayed atop the blank template (black) to demonstrate an imperfect alignment}
    \label{fig:alignment_results}
\end{figure}

For the \textit{comprehensive} format, we verify alignment using a homography matrix computed from the filtered matches from \ref{sec:methods-segmentation}. An image is considered successfully aligned if we exceed a 15 match threshold, restricting to only those matches with a maximum re-projection error of 6.0 pixels. Otherwise, we increase ORB match pairs (5,000 $\rightarrow$ 7,000 $\rightarrow$ 10,000) across three attempts. If all three attempts fail, the image is flagged for manual inspection. This method achieves a 99.7\% success rate on 836 randomly sampled property assessment cards.  The two misaligned cases were likely due to poor scanning. We show an Example image alignment in Figure \ref{fig:alignment_results}; this scan was chosen to show that even an imperfect alignment can yield high fidelity for the table cell contents.

For the \textit{single-cell} format, there are two metrics of interest when evaluating the segmentation method: the success rate of extracting a segment and the accuracy of extracting the correct segment. For the success rate, we use our segmentation algorithm on 56,037 documents and are able to successfully extract segments for 49,845 of them -- i.e., a rate of 89.0\%. To evaluate the accuracy, we randomly sample 499 assessment cards and examine the tables to compare if the extracted table segment is correct. We find only 1 error case where the segment represents the second cell in the column instead of the first, giving an accuracy of 99.8\%.

\subsubsection{OCR Model Results}
\label{sec:results-ocrmodels}

\begin{table}[]
\centering
\resizebox{\columnwidth}{!}{%
\begin{tabular}{|c|cccc|c|}
\hline
\multirow{2}{*}{Metrics} & \multicolumn{4}{c|}{Single-cell} & Comprehensive \\ \cline{2-6}
                         & Top 90\% & Top 95\% & Top 99\% & All 100\% & All 100\% \\ \hline
\begin{tabular}[c]{@{}c@{}}$R^2$\\ (higher is better)\end{tabular}         & 0.77& 0.70& 0.64& 0.63& 0.76     \\ \hline
\begin{tabular}[c]{@{}c@{}}MAPE \\ (lower is better)\end{tabular}                         & 5.42\%& 10.36\%  & 13.86\%  & 14.72\%   & 3.25\%   \\ \hline
\begin{tabular}[c]{@{}c@{}}RMSPE \\ (lower is better)\end{tabular}                        & 26.21\%  & 34.26\%  & 38.97\%  & 40.04\%   & 36.59\%  \\ \hline
\begin{tabular}[c]{@{}c@{}}MPE \\ (lower is better)\end{tabular}                          & 0\%      & 0\%      & 0\%      & 0\%       & 0.36\%   \\ \hline
\begin{tabular}[c]{@{}c@{}}Within 5\% \\ of True Value \\ (higher is better)\end{tabular} & 94.68\%  & 89.73\%  & 84.19\%  & 85.37\%   & 96.52\%  \\ \hline
\begin{tabular}[c]{@{}c@{}}Within 10\% \\ of True Value\\ (higher is better)\end{tabular} & 94.71\%  & 89.76\%  & 86.25\%  & 85.39\%   & 96.98\%  \\ \hline
\begin{tabular}[c]{@{}c@{}}Within 20\% \\ of True Value\\ (higher is better)\end{tabular} & 94.72\%  & 89.77\%  & 86.26\%  & 85.40\%   & 97.38\%  \\ \hline
\end{tabular}%
}
\caption{Prediction metrics of OCR models for different confidence thresholds, with comprehensive and single-cell format results}
\label{tab:ocr-metrics}
\end{table}

For the \textit{single-cell} format, we find the best performing OCR model to be TrOCR fine tuned on a mixture of our Hamilton county dataset combined with additional handwritten digit data from CAR-B. Fine tuning on the DIDA dataset was found to be detrimental since the digit strings are primarily year values recorded in church documents, causing the fine tuned TrOCR model to incorrectly predict values between 1800-1940 more often. The results we present are from the best performing TrOCR model, trained on 7375 randomly sampled entries from our Hamilton county dataset and 3000 entries from CAR-B. We find that this model is relatively accurate with low MAPE and MPE values. Howeve, while errors are rare, as evidenced by 85\% of all predictions falling within 5\% of their true values, the magnitude of the errors are large. This is often due to the insertion or deletion of digits, which can create errors that are orders of magnitudes off.

We note that another common error case is where TrOCR fails to detect recognizable digits. In this case, it will output a blank prediction which is converted to a prediction value of "0" for the purpose of our analysis. Fortunately, these errors are usually accompanied by a low confidence score which allows these low confidence predictions to be filtered. By choosing an appropriate threshold, we can achieve an exact match accuracy of up to 99.4\%. To evaluate the impact of this filtering on the outputs of the model, we report the accuracy metrics for retaining top 90\%, 95\% and 99\% of the most confident predictions, see Table \ref{tab:ocr-metrics}. We see significant improvements in the model performance if we retain only the top 90\% of the most confident predictions, achieving a MAPE of 5.42\% and predictions within 5\% of the true value for 94.68\% of the test cases.

For the \textit{comprehensive} format, where the OCR model evaluates all numerical cells in an image, we observe better overall accuracy compared to the single-cell setting. This model achieves an $R^2$ score of 0.76 compared to 0.63 for the \textit{single-cell} format when retaining all confidence values. It also demonstrates lower MAPE (3.25\% vs. 14.72\%) and RMSPE (36.59\% vs. 40.04\%), indicating reduced overall error. Additionally, the percentage of predictions within 5\% of the true value is higher, reaching 96.52\% compared to 85.37\% for the \textit{single-cell} format. These differences are largely attributable to the fact that many of the numbers predicted in the comprehensive setting are two-digit years, which are easier to identify than the three-to-five digit valuations of the \textit{single-cell} format.

\section{Predicting Historical Home Values}
\label{sec:regression-models-for-housing}

In an ideal scenario, we would have the scanned images of all historical property assessments for every property in the United States. Unfortunately, many local governments have not had the resources or ability to scan their records. Hiring personnel to travel to local governments around the country and scan their historical records would be extremely expensive and time consuming. Thus, in the absence of scanned historical records, we propose a predictive model that can estimate the historical assessment.

As discussed in \ref{introduction}, prior to the National Housing Act of 1934, government assessors determined property values based on their construction characteristics. Many assessors used the same manuals that provided equations for estimating home assessments based on know features. Therefore, our aim is to reverse engineer these equations by learning a regression model to predict the relationship between property features and their historical assessment values (\S \ref{sec:proposed-regression-model}).
Because our model is trained on data from only one county (Hamilton), we then turn to the question of whether it can generalize to properties from a different county (Franklin) (\S \ref{sec:testing-generalizability}).
We also evaluate the model for algorithmic bias by examining its performance across communities of varying demographic compositions (\S \ref{sec:checking-for-bias}) and consider the cost-accuracy tradeoffs associated with our proposed solutions (\S \ref{sec:cost-acc-tradeoff}).

\subsection{Related Work}
\label{sec:regression-related-work}

Using building features to predict home values has a long history within the real estate profession. As just outlined, this was the basis for the original assessment values. They tabulated charts, rather than using machine learning models, but the original assessment calculations were at their core predictions of home value based on property features.

Although the federal government's introduction of racialized appraisal methods phased out the use of construction cost-based models, the industry continued to use predictive models to estimate property values. Yet, these models began to reflect the new racialized approaches of appraising as implemented by the sales comparison approach. Today, several companies and scholars design Automated Valuation Models (AVMs) that attempt to use available property and sales data to emulate the appraiser's sales comparison methodology\cite{zhao2019deep, ho2021predicting, trawinski2017comparison, baldominos2018identifying, tchuente2022real}. Similar to this work, our models are using property features to predict value. However, unlike contemporary AVM models, we are not trying to predict value based on recent sale data. Instead, we are attempting to retrospectively impute historical assessments based on the historical construction cost. To our knowledge, no one else has attempted this particular task.

\subsection{Regression Model}
\label{sec:proposed-regression-model}

\subsubsection{Property Feature Data}
\label{sec:feature_data}

To link the historical assessment values to building features, we use each parcel's contemporary property record which includes detailed information about a home's size, characteristics, and construction quality. Although some properties have undergone major revisions that have increased the building square footage or number of rooms, the vast majority of properties built before 1930 remain the same size--enabling us to use contemporary features as a rough approximation of basic property characteristics. See Table \ref{table:features} for the full list of features we employ in the models. Categorical features are given a one-hot encoding. In Appendix \ref{cleaningappendix}, we detail how we processed the data for our analysis.

\begin{table}[t]
\centering
\begin{tabular}{|l|p{0.5\linewidth}|}
\hline
Square footage & attic, basement, floor 1, floor 2, half-floor, total livable area \\
\hline
Building characteristics & stories, style, grade/condition of building, exterior wall type, basement type, heating type, air conditioning type, total number of rooms, total full and half bathrooms, number of fireplaces, garage type and capacity \\
\hline
Parcel characteristics & land use code, neighborhood, number of sub-parcels \\
\hline
\end{tabular}
\caption{Features built from contemporary data}
\label{table:features}
\end{table}

\subsubsection{Methods}
\label{sec:methods-regression}

We formulate the task of predicting a historical value from contemporary property data as a standard regression problem where the target value to be predicted is the labeled 1933 building assessment value. As mentioned in Section \ref{sec:manual-annotation}, we have 10,452 parcels in Hamilton County with labels collected by hand, which we merge with the contemporary feature data outlined in Table \ref{table:features} to create the training and test matrices. We use an 80\%-20\% train-test split, and employ 5-fold cross validation within the training set for hyperparameter tuning.

We use a stepwise approach to model selection. First, using a single set of default hyperparameters, we train many different model classes and observe performance on a validation set. The results of this exercise are in Appendix \ref{app:model selection}. We then select the best performing model classes, and conduct a more extensive hyperparameter grid search, selecting the best model using the 5-fold cross validation root mean squared error (RMSE).

This approach leads us to choose a random forest regressor, a non-linear ensemble of decision tree regressors,  as the best model. The hyperparameters of this model are in Table \ref{table:chosen ml model}.

\begin{table}[t]
\centering
\begin{tabular}{|l|c|}
\hline
Model class & Random forest regressor\\
\hline
Number of estimators & 2500\\
\hline
Max depth & 200\\
\hline
Minimum samples for split & 4\\
\hline
Max features & sqrt\\
\hline
\end{tabular}
\caption{Chosen Regression model}
\label{table:chosen ml model}
\end{table}

While OCR methods and regression methods can be viewed as two separate approaches for predicting the same target variable, they accomplish their task using different inputs and techniques. These two methods are complementary to each other in that they can be combined in various ways to improve performance. In this work, we use the trained OCR model to create annotated labels for training the regression model. This allows the use of all 56,037 retrieved scanned documents for training and testing of the regression model instead of only the 10,452 manually labeled samples. We show in Section \ref{sec:results-ml} that this improves the performance of the regression models.

\subsubsection{Metrics}
\label{sec:experimental-details}
We use several common statistical evaluation metrics for regression tasks including coefficient of determination ($R^2$), Mean Absolute Error (MAE), Mean Absolute Percentage Error (MAPE), Root Mean Squared Percentage Error (RMSPE), the Median Percentage Error (MPE), and the percentage of test cases where we predicted a value that is within 5\%,  10\%, or 20\% of the true value. We report results on buildings in the middle 90\% of properties based on appraised value (i.e., 5th to 95th percentile), to reduce the effect of outliers. These outliers could either reflect data entry errors at the time, or very expensive properties that are unlikely to be of core interest to researchers studying the general effects of redlining.

\begin{table}[t]
\centering
\resizebox{\columnwidth}{!}{%
\begin{tabular}{|c|c|c|c|c|}
\hline
Metrics                                                                                   & OCR& Regression & Augmented& Generalization\\ \hline
\begin{tabular}[c]{@{}c@{}}$R^2$ \\ (higher is better)\end{tabular}                   & 0.63&  0.62& 0.74& 0.38\\ \hline
\begin{tabular}[c]{@{}c@{}}MAE \\ (lower is better)\end{tabular}                          & \$492     &  \$489  & \$452     & \$571 \\ \hline
\begin{tabular}[c]{@{}c@{}}MAPE \\ (lower is better)\end{tabular}                          & 14.72\%     &  17.48\%  & 16.12\%     & 22.72\% \\ \hline
\begin{tabular}[c]{@{}c@{}}RMSPE \\ (lower is better)\end{tabular}                         & 40.04\%    &  27.73\%  & 24.01\% & 38.71\% \\ \hline
\begin{tabular}[c]{@{}c@{}}MPE \\ (lower is better)\end{tabular}                 & 0\%       &  10.60\%  & 11.27\%    & 28.31\% \\ \hline
\begin{tabular}[c]{@{}c@{}}Within 5\% \\ of True Value \\ (higher is better)\end{tabular} & 85.36\% &  25.81\%  & 24.39\% & 15.70\% \\ \hline
\begin{tabular}[c]{@{}c@{}}Within 10\% \\ of True Value\\ (higher is better)\end{tabular} & 85.39\% &  48.06\%  & 45.85\% & 31.40\% \\ \hline
\begin{tabular}[c]{@{}c@{}}Within 20\% \\ of True Value\\ (higher is better)\end{tabular} & 85.40\% &  73.44\%  & 74.55\% & 62.50\% \\ \hline
\end{tabular}%
}
\caption{Prediction performance of evaluated models}
\label{tab:model-metrics}
\end{table}

\subsubsection{Results}
\label{sec:results-ml}

\begin{figure}[t]
    \centering
    \includegraphics[width=\columnwidth]{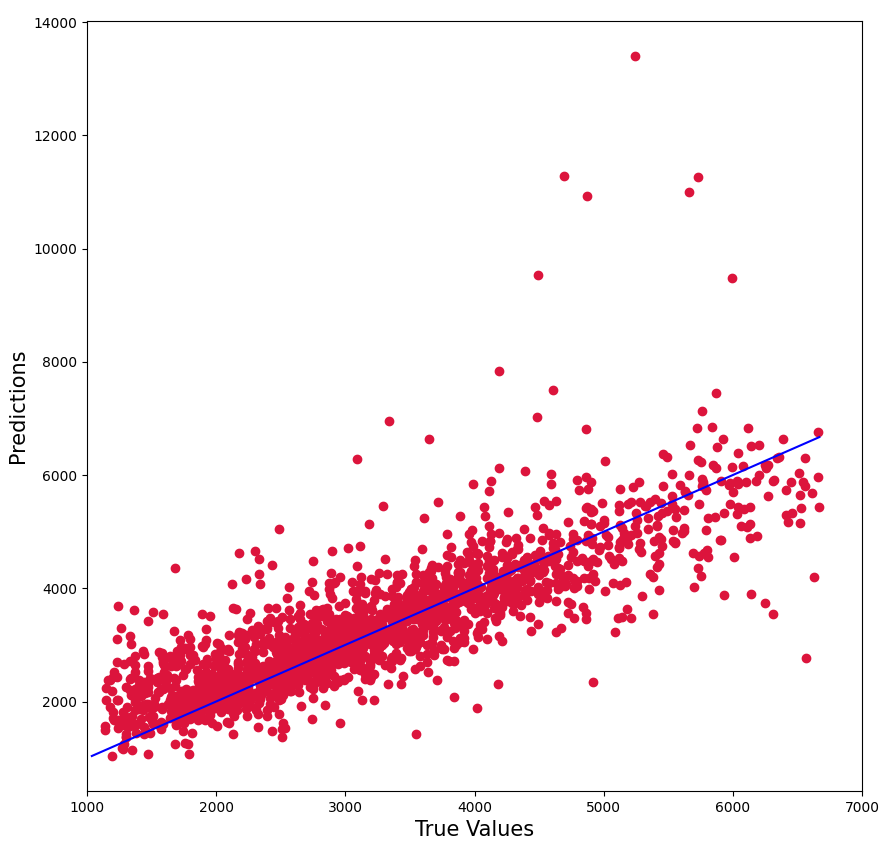}
    \centering
    \caption{Regression Model Predictions}
    \label{fig:ml-model}
\end{figure}

The statistics of the best performing models from our experiments are shown in Table \ref{tab:model-metrics}.
The chosen random forest regressor model predicts the target value with an MAPE of 17.48\%. As seen in Figure \ref{fig:ml-model}, the model seems to perform worse on higher-value properties, with larger over-predictions and under-predictions. Many of the square footage-related features and other building characteristics such as grade, wall type, and number of rooms are in the top 10 most important features based on impurity reduction. See Appendix \ref{app:feature importances} for a plot of the feature importances.

\begin{figure}[t]
    \centering
    \includegraphics[width=\columnwidth]{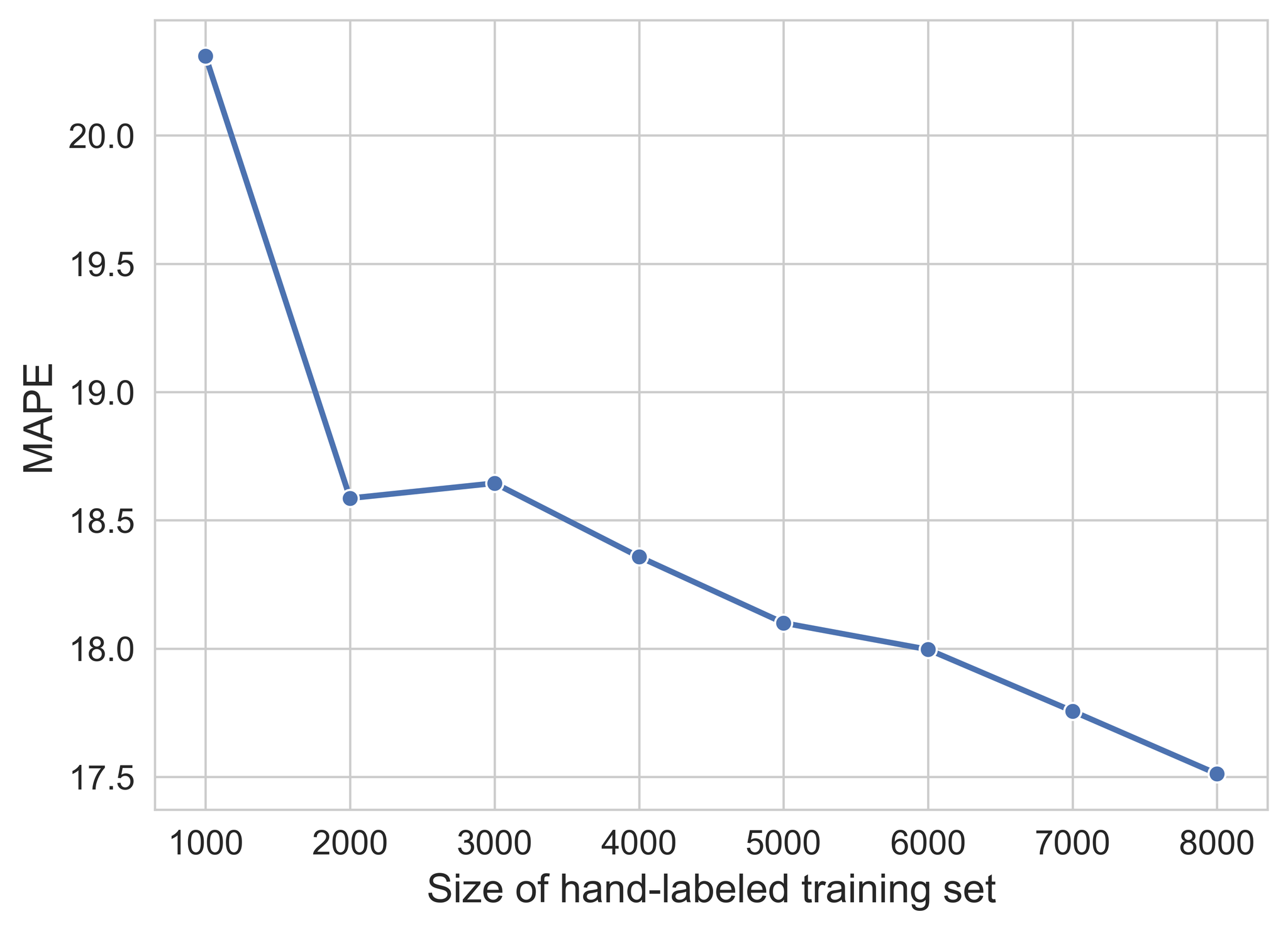}
    \centering
    \caption{MAPE as size of hand-labeled training data increases}
    \label{fig:mape_v_n_ml_handonly}
\end{figure}

A relevant question for our proposed approach is the number of samples that need to be manually digitized for the model to predict the target value accurately. Figure \ref{fig:mape_v_n_ml_handonly} shows the improvement in MAPE as the size of the training set increases. As the number of labeled samples in the training set increases from 3,000 to 8,000, the MAPE drops from roughly 18.6\% to 17.5\%. We did not collect additional samples, but based on the trend it appears that additional data would improve performance.

\paragraph{Training on OCR-labeled Data (`Augmented' regression model)}

\begin{figure}[t]
    \centering
    \includegraphics[width=\columnwidth]{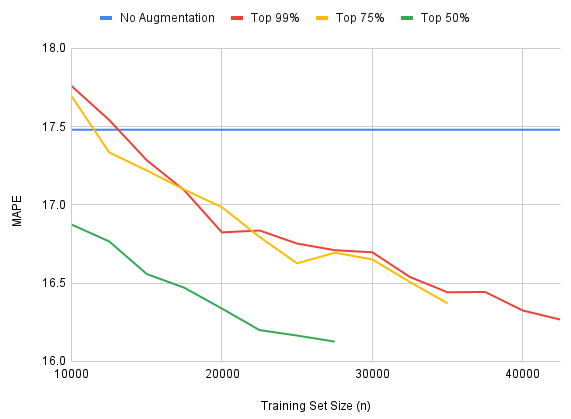}
    \centering
    \caption{MAPE and OCR Confidence Threshold vs n}
    \label{fig:ocr_threshold}
\end{figure}

Next we incorporate the OCR-labeled 1933 assessment values as training data.
By including only OCR labels with confidence above some threshold, we can trade off between quantity and quality of the OCR training samples.
To examine this effect, the performance of the augmented regression models using different OCR prediction confidence thresholds retaining the top 99\%, 90\% 75\% and 50\% of the most confident predictions, is shown in Figure \ref{fig:ocr_threshold}. Compared to the regression model performance listed in Table \ref{tab:model-metrics} while we see a slight improvement in some accuracy measures such as MAPE from 17.48\% to 16.12\% and RMSPE from 27.73\% to 24.01\%, other measures such as MPE see a slight decline. We also note that while the number of predictions within 20\% of the true values improved from 73.44\% to 74.55\%, the predictions within 5\% and 10\% of the true values decreased. This result suggests that it is not conclusive that augmenting the regression models using OCR predictions is beneficial. From our analysis, the outliers in the OCR predictions, despite our efforts to remove them by applying a threshold for the OCR prediction confidence, are highly detrimental to regression models and offset the benefits of additional training samples.

\paragraph{Hamilton County where OCR Methods Failed}

We also test our regression model's performance on the group of 6,192 Hamilton County parcels for which our OCR methods failed during segmentation. We manually labeled a random sample of 778 of these cases and evaluate our augmented regression model's predictions on these samples to determine whether the model's performance on this subset is similar to those from our other experiments.
The results of our model on these OCR segmentation failures is shown in Table \ref{tab:ocr-failures}. We observe no substantial difference in prediction performance, which indicates that our training sample is not problematically biased by earlier failures in the pipeline.

\begin{table}[]
\centering
\begin{tabular}{|c|c|c|}
\hline
Metrics                                                                                   & Augmented & OCR Failures \\ \hline
\begin{tabular}[c]{@{}c@{}}R\textasciicircum{}2\\ (higher is better)\end{tabular}         & 0.74& 0.67\\ \hline
\begin{tabular}[c]{@{}c@{}}MAPE \\ (lower is better)\end{tabular}                         & 16.12\%      & 15.98\%      \\ \hline
\begin{tabular}[c]{@{}c@{}}RMSPE \\ (lower is better)\end{tabular}                        & 24.01\%      & 25.18\%      \\ \hline
\begin{tabular}[c]{@{}c@{}}MPE \\ (lower is better)\end{tabular}                          & 11.27\%      & 11.56\%      \\ \hline
\begin{tabular}[c]{@{}c@{}}Within 5\% \\ of True Value \\ (higher is better)\end{tabular} & 24.39\%      & 23.68\%      \\ \hline
\begin{tabular}[c]{@{}c@{}}Within 10\% \\ of True Value\\ (higher is better)\end{tabular} & 45.85\%      & 45.82\%      \\ \hline
\begin{tabular}[c]{@{}c@{}}Within 20\% \\ of True Value\\ (higher is better)\end{tabular} & 74.55\%      & 73.79\%      \\ \hline
\end{tabular}
\caption{Regression Model Generalization on OCR Failures}
\label{tab:ocr-failures}
\end{table}

\subsection{Testing Generalizability}
\label{sec:testing-generalizability}

\subsubsection{Franklin County Historical Assessments}

To test whether our regression model is generalizable to other counties, we selected a second county that has publicly available scanned historical property records: Franklin County (Columbus), Ohio\footnote{The data is available on the county auditor website: \url{https://apps.franklincountyauditor.com/Outside_User_Files/2023/}}. Although slightly different in structure and years assessed, Franklin County's records are similar in form to Hamilton County.
We manually annotated a randomly drawn subset of 506 cards from Franklin county for use as a test set.

\subsubsection{Method}
\label{sec:generalization}

To apply the trained regression model to make predictions in Franklin County, we had to ensure that the features in Franklin County were comparable to those used in the Hamilton model. While some of the important features were common (e.g., square footage of floor 1), several features were not available in Franklin County or were captured in a different format (e.g. presence of attic rather than its square footage). To test generalization, we train the model with only the subset of features that were comparable across both counties, and use this limited model to report performance on the Franklin County test set. See Appendix \ref{app:hamilton franklin features} for more details on the feature subset used.

\subsubsection{Results}
\label{sec:results-generalization}

\begin{figure}[t]
    \centering
    \includegraphics[width=\columnwidth]{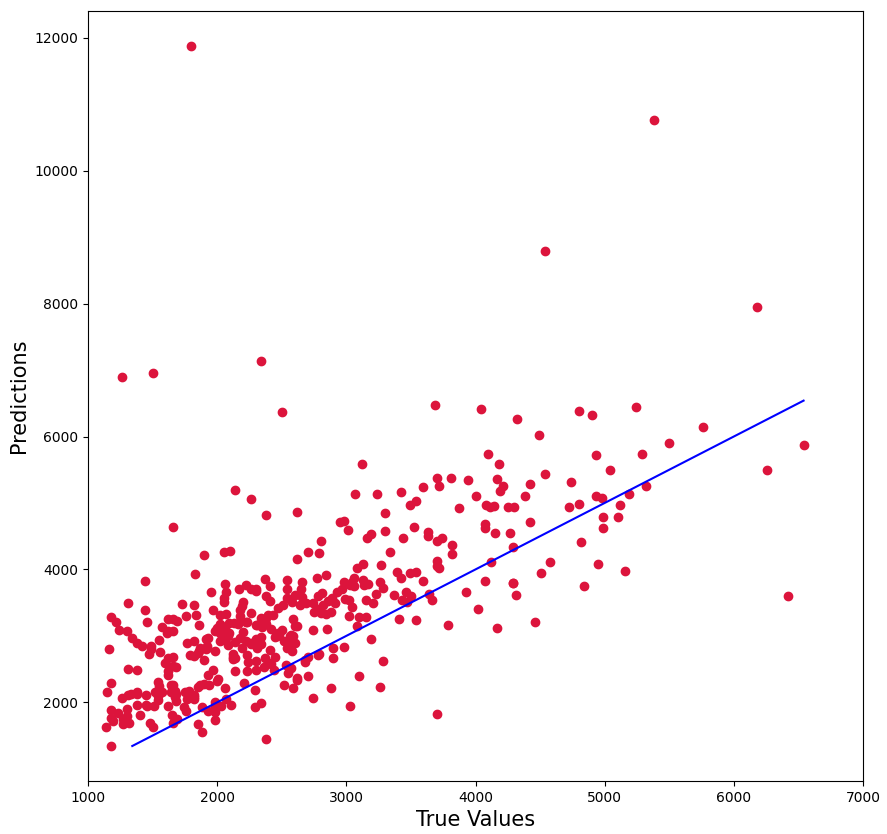}
    \centering
    \caption{Regression Model Predictions on Franklin County}
    \label{fig:franklin}
\end{figure}

We observe that the distributions of the target values of the two counties are different with the median target value being \$2,300 in Franklin County, which is lower than the Hamilton County median of \$3,085. To correct for the difference between these two distributions we randomly sample 100 parcels in Franklin County and compute the mean and standard deviation of the two counties, applying the adjustment in Equation \ref{eq:adjustment}.
\begin{equation} \label{eq:adjustment}
Y_{Franklin}
= \frac{Y_{Hamilton} - \mu_{Hamilton}}{\sigma_{Hamilton}}*\sigma_{Franklin} + \mu_{Franklin}
\end{equation}
The results for Franklin County are worse than those for the Hamilton County test set, with an MAPE of 22.72\% (see  Table \ref{tab:model-metrics}). Figure \ref{fig:franklin} shows that the model predictions correlate less well with the true values. This suggests that to generalize fully across multiple U.S. counties, training data from multiple counties may be needed so the model can better learn regional differences in construction costs.

\subsection{Checking for Bias in Model Predictions}
\label{sec:checking-for-bias}

Given that social scientists interested in this historical data are primarily interested in the ways federal policies have influenced racial and socioeconomic inequality, it is imperative to ensure our models are performing comparably across communities of different demographic compositions.

\subsubsection{Data}
We use 2020 U.S. American Community Survey 5 year summary files to gather census tract data on median income, poverty, racial composition, owner occupacy, and single-family homes. We identify which census tract each parcel falls in by combining parcel footprints (i.e., polygons identifying parcel boundaries) from the Cincinnati Area GIS portal\footnote{Open data sourced from \url{https://data-cagisportal.opendata.arcgis.com/}} with census tract boundaries.

\subsubsection{Results}
Our test set, on which we report all previous performance statistics, spans 160 different census tracts. These tracts have sufficient variation in the key variables for our bias analysis: median annual income ranges from \$11,831 to \$161,964 and the proportion of White people in the population ranges from 6\% to 98\%.

\begin{figure}[t]
    \centering  \includegraphics[width=\columnwidth]{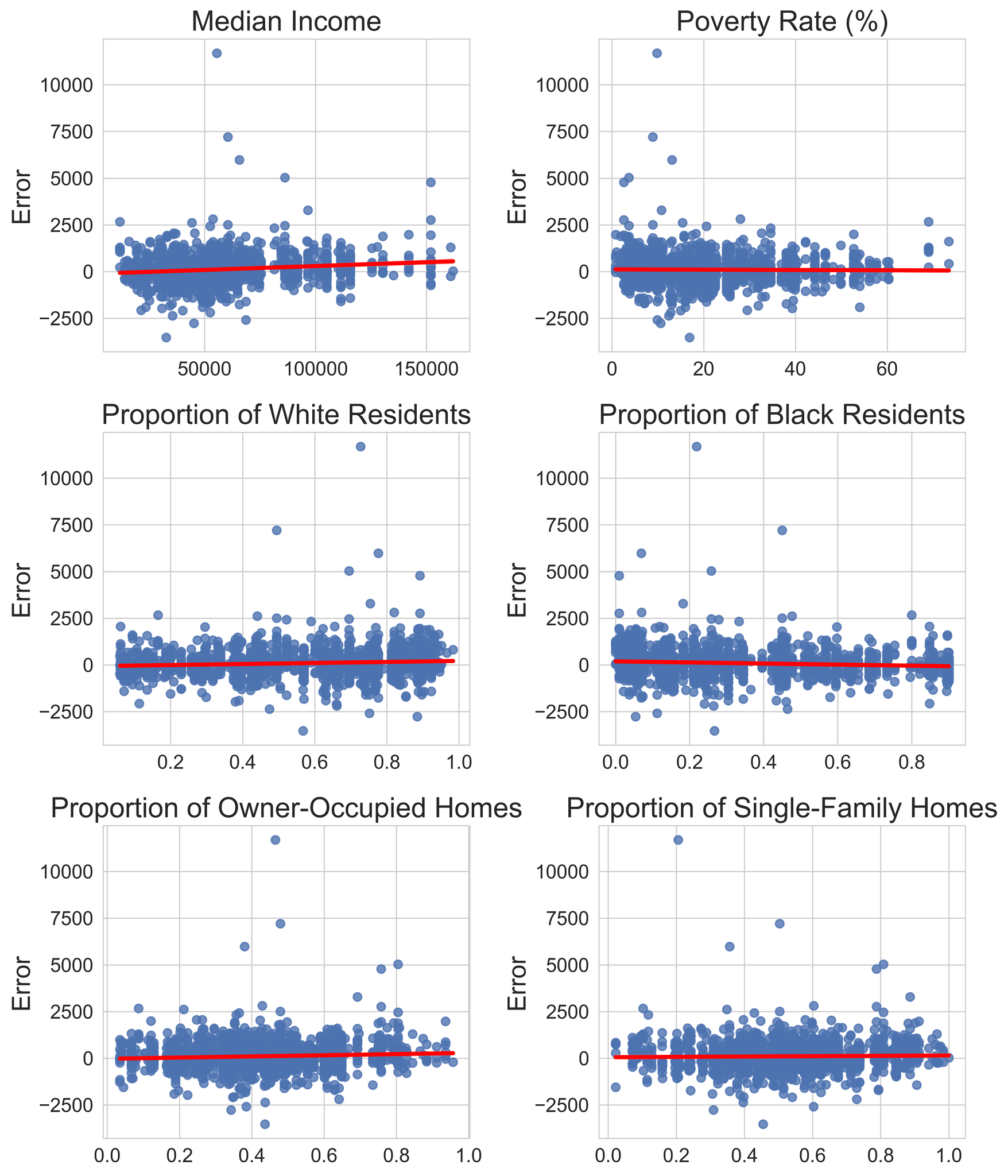}
    \centering
    \caption{Error (Predicted Value $-$ Actual Value) against Key Socio-Demographic Characteristics}
    \label{fig:bias-scatter}
\end{figure}

Figure \ref{fig:bias-scatter} shows a series of plots of model error on each test observation (i.e., predicted value $-$ actual value) against income, race and housing-related variables at the tract-level. There are no systematic patterns in the observed errors. Table \ref{tab:bias-corr-table} reports the pairwise correlation of Absolute Percentage Error with the demographics variables of interest. We see that the correlations range from $-0.068$ to $0.070$, consistent with the scatter plots.

\begin{figure}[t]
    \centering  \includegraphics[width=\columnwidth]{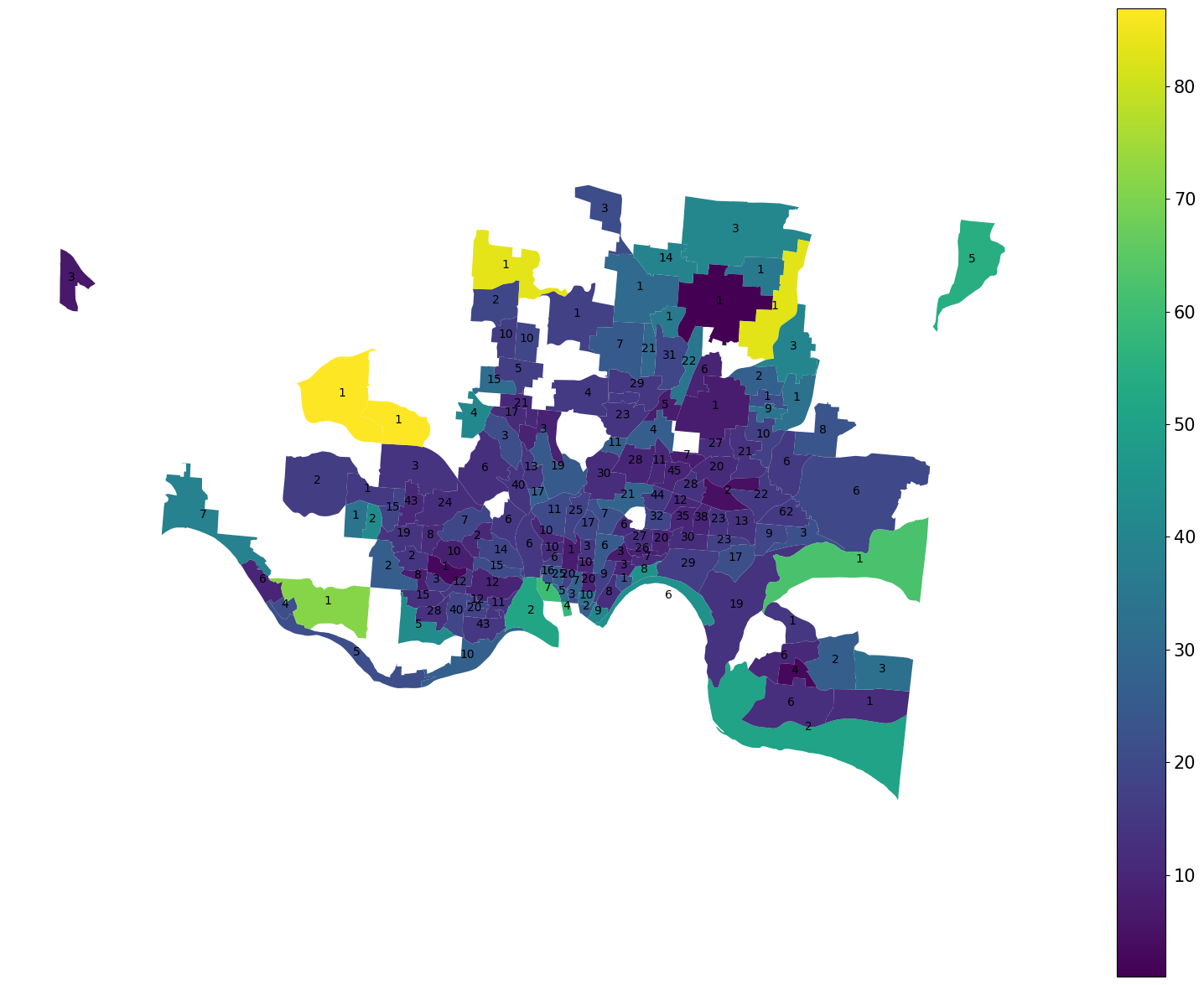}
    \centering
    \caption{Heatmap of MAPE by census tract---numbers indicate the count of test samples in that tract}
    \label{fig:bias-map}
\end{figure}

We also check to see if there is any spatial pattern in the model errors. Figure \ref{fig:bias-map} shows a map with each census tract colored by the mean absolute percentage error of all test observations in the tract. The tracts with notably higher errors (in yellow and light green on the outskirts of Cincinnati) are tracts with very few test samples; for the rest of the tracts with similar numbers of test samples, there seem to be no clear spatial patterns in errors.

\begin{table}[]
\centering
\begin{tabular}{|c|c|}
\hline
Variable & Correlation \\
\midrule
Proportion LatinX Population & $-0.068$ \\
Proportion Black Population & $-0.045$ \\
Total Population & $-0.044$ \\
Proportion Race (Other) & $-0.005$ \\
Proportion Vacant Lots & $0.006$ \\
Poverty Rate & $0.023$ \\
Proportion Asian Population & $0.026$ \\
Proportion Indigenous Population & $0.045$ \\
Proportion Single-Family Housing & $0.052$ \\
Proportion White Population & $0.053$ \\
Proportion Owner Occupied Housing & $0.053$ \\
Median Income & $0.070$ \\
\bottomrule
\end{tabular}%
\caption{Correlation between Absolute Percentage Error and Socio-Demographic Variables }
\label{tab:bias-corr-table}
\end{table}

\subsection{Cost/Accuracy Trade-off}
\label{sec:cost-acc-tradeoff}

One of the main benefits of the proposed OCR and regression techniques for extracting values from historical records is the ability to scale to large numbers of documents with minimal cost. As a baseline we consider a hypothetical collection of 353,973 cards, each with a single value to extract (i.e., the same number of properties in Hamilton County). The cost and accuracy comparison of the two proposed methods is shown in Figure \ref{fig:cost-comparison}. For details on the following estimates calculations, see Appendix \ref{app:costs}.

\begin{figure}[t]
    \centering
    \includegraphics[width=\columnwidth]{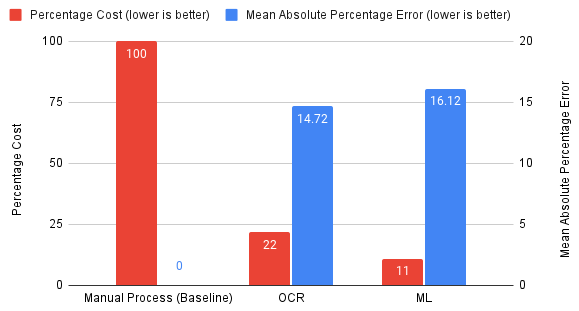}
    \centering
    \caption{Cost and Accuracy Comparisons of Proposed Methods}
    \label{fig:cost-comparison}
\end{figure}

To estimate the cost savings of the OCR methods, we assume scanned documents are available and compare the estimated costs of manual data entry of a single target value against adapting the OCR methods. Manually extracting a single value from scanned documents at the rate we used for the manual labeling process on 353,973 documents will cost an estimated \$24,789.22. By contrast, the cost of employing a data scientist to adapt the OCR methods described in this work to a different document will cost \$5,568.10, which is 22\% of the manual process. The drawback for this cost reduction is the reduction in accuracy with an MAPE of 14.72\%. Based on our experience with hand-labeling, given the structured nature of these documents, we assume that manual collection would yield close to perfect accuracy if clear instructions are provided and the work is well distributed. This assumption may not hold if the quality control of manual collection is difficult, and thus reduces the relative accuracy cost of OCR.

In the scenario where scanned documents are not available, we compared the cost of manual scanning and data entry against the proposed regression methods. Using online estimates of document scanning services, this will incur an average cost of \$35,570.42 for 353,973 documents. Combined with the data entry costs listed previously, this gives a total cost of \$60,359.64 for the manual process. We estimate that the cost to develop the regression model described in this work, including the costs of generating 12,423 training samples, to be \$6,816.49. This represents a 11\% of the cost of a comparable manual process but using this method further reduces the accuracy to an MAPE of 17.48\%. There is a cost-accuracy tradeoff even within the regression method: as shown in Figure \ref{fig:mape_v_n_ml_handonly}, one could incur a higher or lower cost of hand-labeling training samples based on the desired accuracy.

\section{Conclusion}

Using our mix of machine learning and computer vision methods, we were able to successfully create the first county-wide dataset of historical property assessment values as well as an initial model for estimating historical assessments based on property features. Our methods were able to predict the values with an accuracy of 14.72\% MAPE and 17.48\% MAPE, respectively. We also demonstrate that these methods are cost effective compared to existing manual methods, saving up to 78\% with the OCR methods and 89\% with regression methods. Though we show the feasibility of augmenting regression model training samples with OCR generated labels, additional work needs to be done to conclusively demonstrate its effectiveness. With potential improvements from expanding the complexity of the regression model, increasing the richness of the building feature inputs, and applying our OCR methods on the full historical document, we expect our proposed methods to perform even better given sufficient time and resources to explore these approaches. This work not only provides a direct service to the social sciences trying to enumerate the impacts of a specific federal policy, but also highlights how machine learning and computer vision can continue to unlock invaluable historical records that can help us study and shape social policies.

\bibliographystyle{ACM-Reference-Format}
\bibliography{citations}

\newpage

\appendix

\section{Testing for Bias from Missing Assessment Cards} \label{app:missing-oc}

We wanted to confirm whether we introduced any bias in our regression models by ignoring the parcels which did not have any assessment cards available. Since we cannot evaluate the model's performance on ground truth values in these cases, we use a Classifier 2 Sample Test \cite{lopezpaz2018revisiting} using the contemporary features to check whether these cases are Missing At Random (MAR) to ensure we do not introduce any bias. We observe a p-value of 0.3870 from the test which confirms that these samples where assessment cards are missing are indeed MAR and does not introduce any bias in our models.

\section{Segmentation} \label{app:segmentation}

\subsection{Our Segmentation Approach (single-cell)} \label{app:segmentation-tesseract}

This task involves recognizing the column header “Buildings” in the image and extracting the bounding boxes of the first cell below it. In this work, we are concerned with extracting the initial construction cost of the building for which we deem the first entry under the "Buildings" column to be a good proxy.

For the task of locating each cell segment, we begin with TesseractOCR as a baseline to label the bounding boxes for sequences of letters and digits. However, this proved to be difficult since there were many false positives and negatives.

\begin{figure}[H]
    \centering
    \includegraphics[width=\columnwidth]{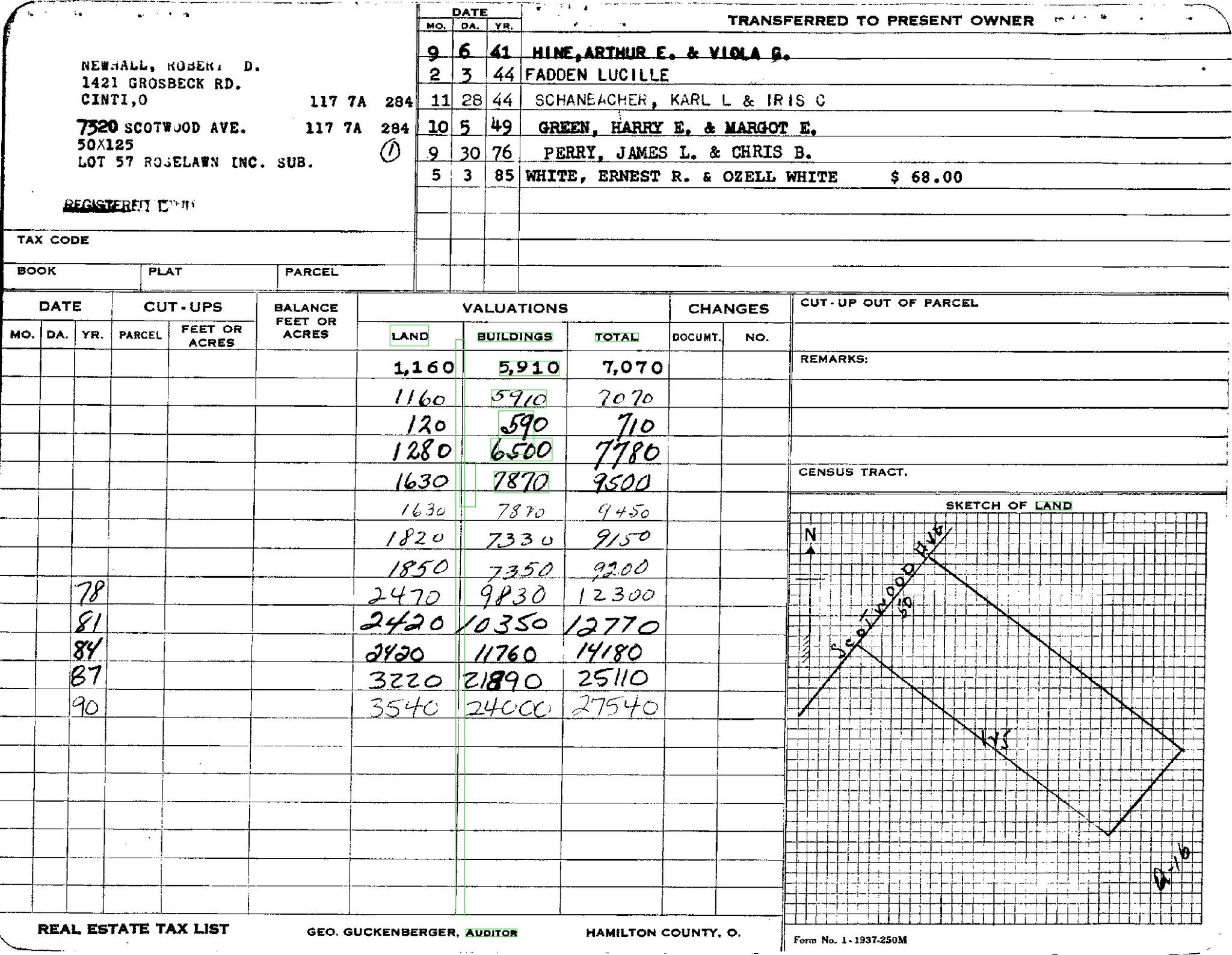}
    \centering
    \caption{Sample TesseractOCR Output}
    \label{fig:sample-tesseract}
\end{figure}

Here we can see several issues. First, there are false positives where non digit elements such grid lines being recognized as characters by TesseractOCR. Second there are false negatives where digits further down the column are not recognized. Furthermore, some sequences of characters are not fully recognized. For example only the "59" of the "590" sequence is recognized. Finally the recognized characters are not always correct. For example, the first three rows were recognized as "5,910", "SULO" and "Ff" of which only the first row is correct. Given that TesseractOCR is a pre-trained model, we found it difficult to modify its behavior for our particular problem and proceeded with building our own multi step solution.

\begin{figure}[H]
    \centering
    \includegraphics[width=\columnwidth]{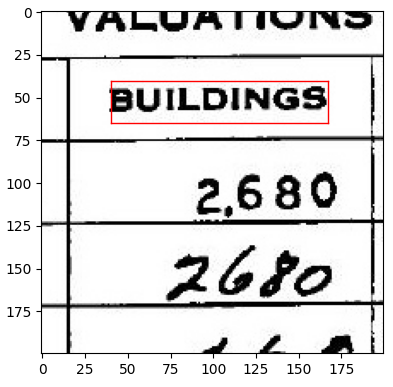}
    \centering
    \caption{Sample cropped document}
    \label{fig:cropped-tesseract}
\end{figure}

For the first step, we retain the use of TesseractOCR for locating the "Buildings" column header and creating a cropped image around the column header. For example of the cropped document containing the detected column header, see Figure \ref{fig:cropped-tesseract}.

\begin{figure}[H]
    \centering
    \includegraphics[width=\columnwidth]{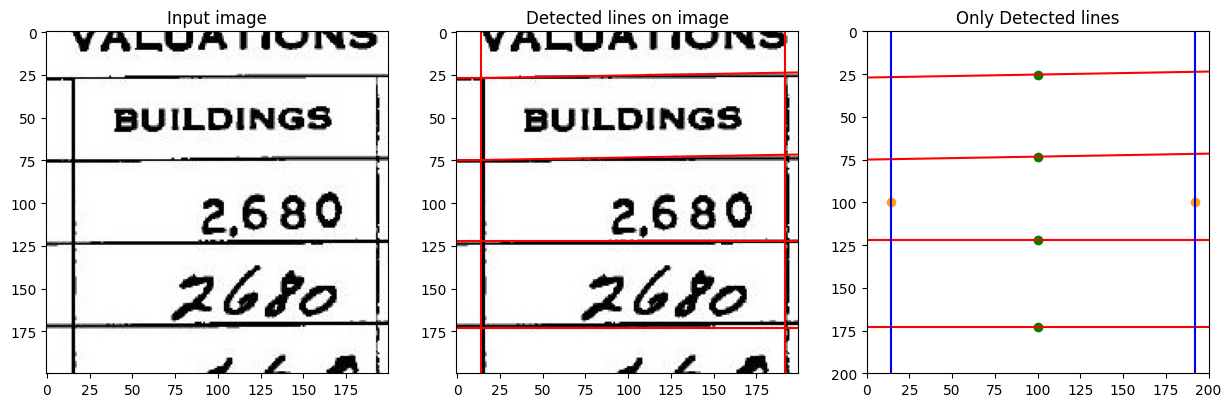}
    \centering
    \caption{Sample line detection using Hough Transform}
    \label{fig:line-detection}
\end{figure}

To extract the cells below the header, we then use Hough Transform \cite{hough_transform} to detect the main line segments in the cropped image. An example of the document with detected lines overlaid on top is shown in Figure \ref{fig:line-detection}.

\begin{figure}[H]
    \centering
    \includegraphics[width=\columnwidth]{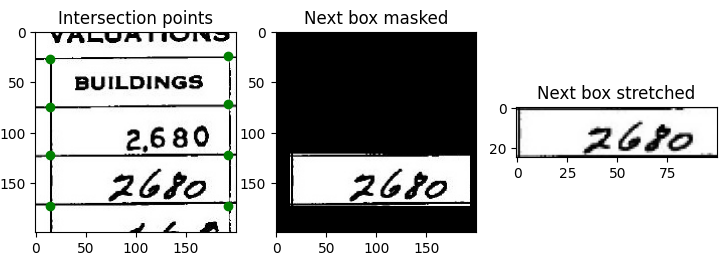}
    \centering
    \caption{Extracting a sample cell as a rectangular image}
    \label{fig:segment-cropped}
\end{figure}

Finally, we use the detected lines and compute the intersections to determine the corners containing the cell we are interested in, which is then used to create a final image of the cell stretched to be a regular rectangle, see Figure \ref{fig:segment-cropped}.

The final output is then ready to be used as an in put to OCR models.

\subsection{LayoutParser Approach} \label{app:segmentation-layoutparser}

\begin{figure}[H]
    \centering
    \includegraphics[width=\columnwidth]{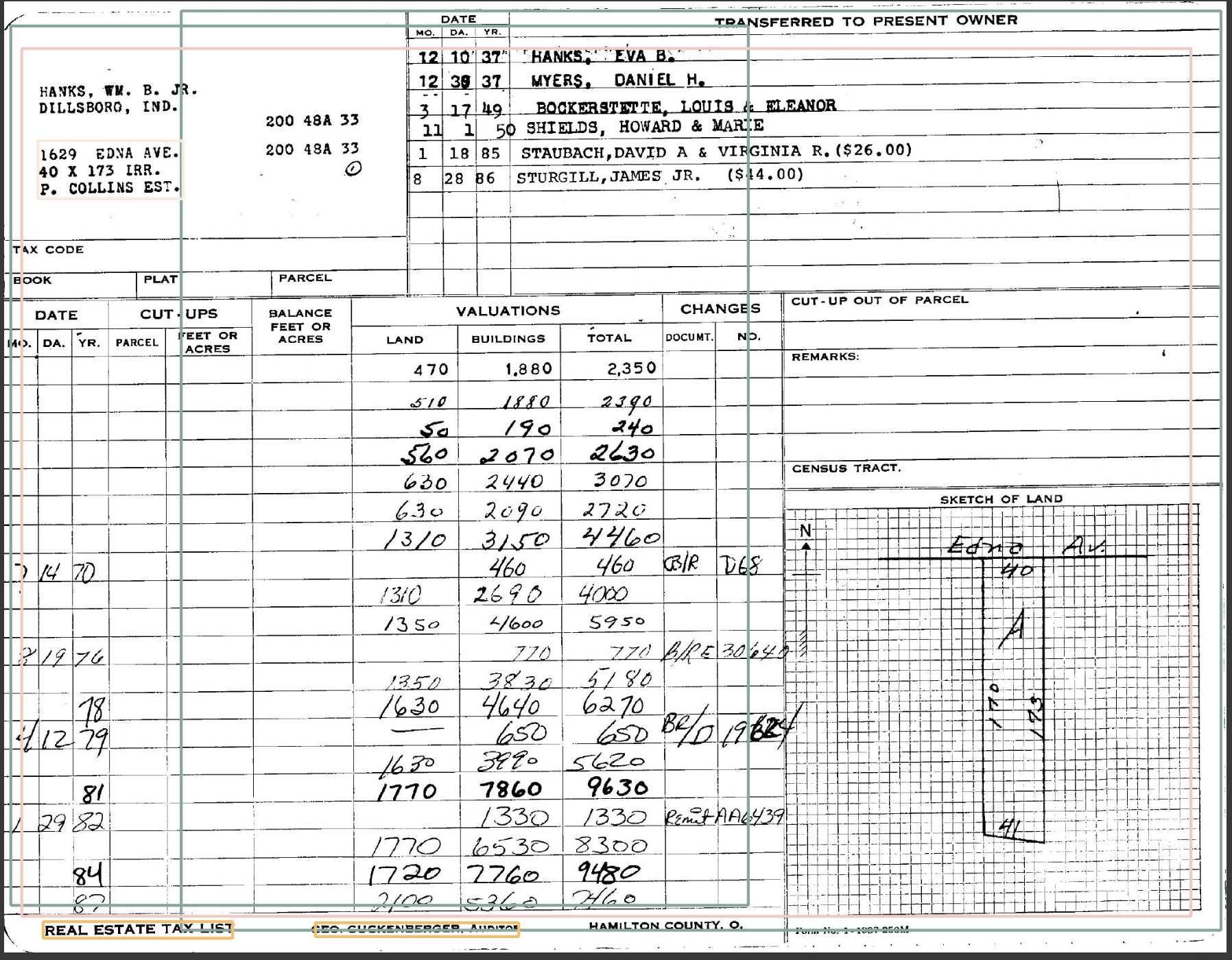}
    \centering
    \caption{Sample LayoutParser Output}
    \label{fig:layout-parser}
\end{figure}

We tried off-the-shelf LayoutParser \cite{shen2021layoutparserunifiedtoolkitdeep} on a subset of our property cards - 10 cards, and it failed to recognize the table cells for all the cards. The results were similar even with different configurations of models used in LayoutParser - MaskRCNN, or FasterRCNN. An example is shown in Figure \ref{fig:layout-parser}.

\subsection{ChatGPT Approach}
\label{app:segmentation-chatgpt}

\begin{figure}[H]
    \centering
    \includegraphics[width=\columnwidth]{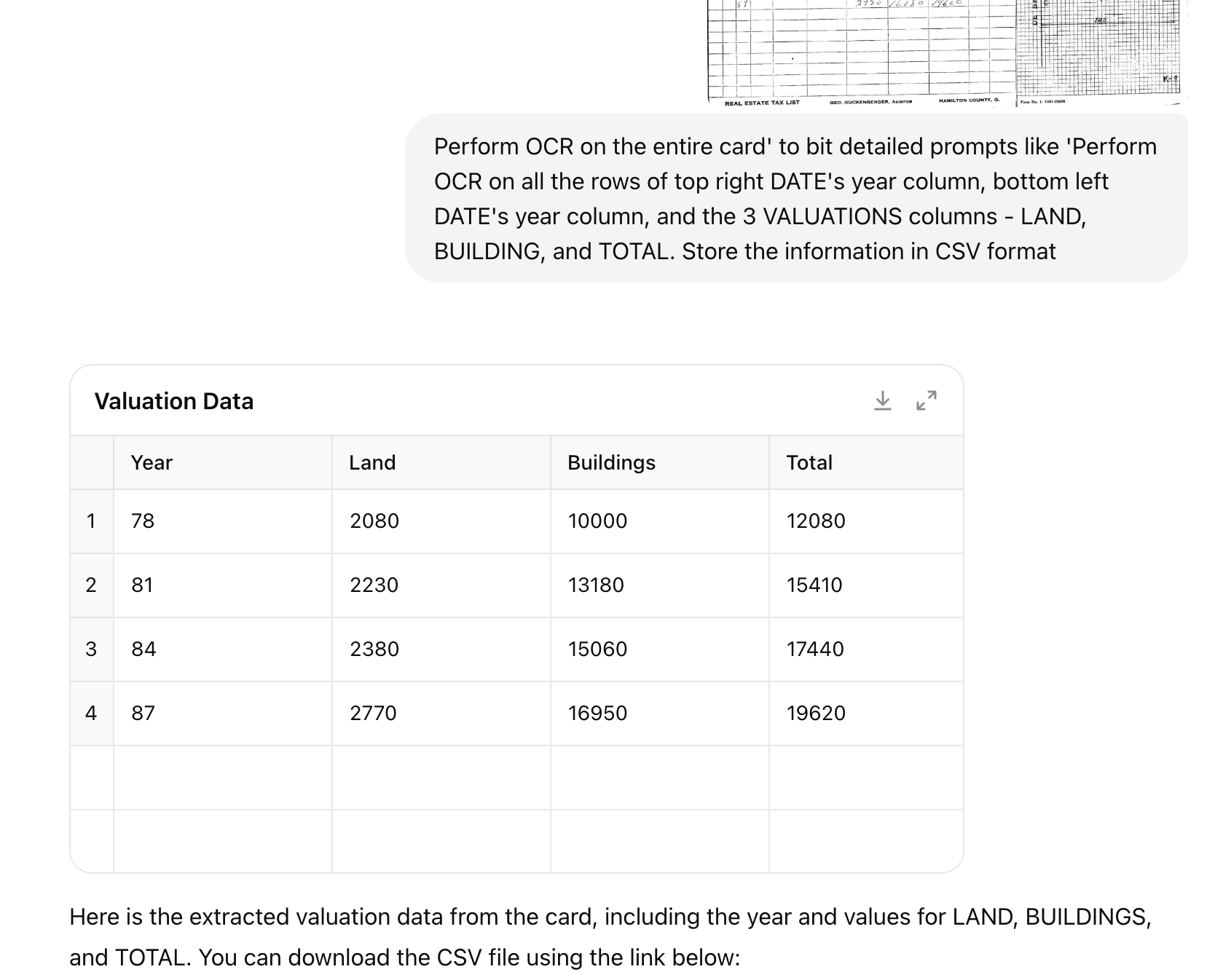}
    \centering
    \caption{GPT returned partial card information}
    \label{fig:gpt2}
\end{figure}

\begin{figure}[H]
    \centering
    \includegraphics[width=\columnwidth]{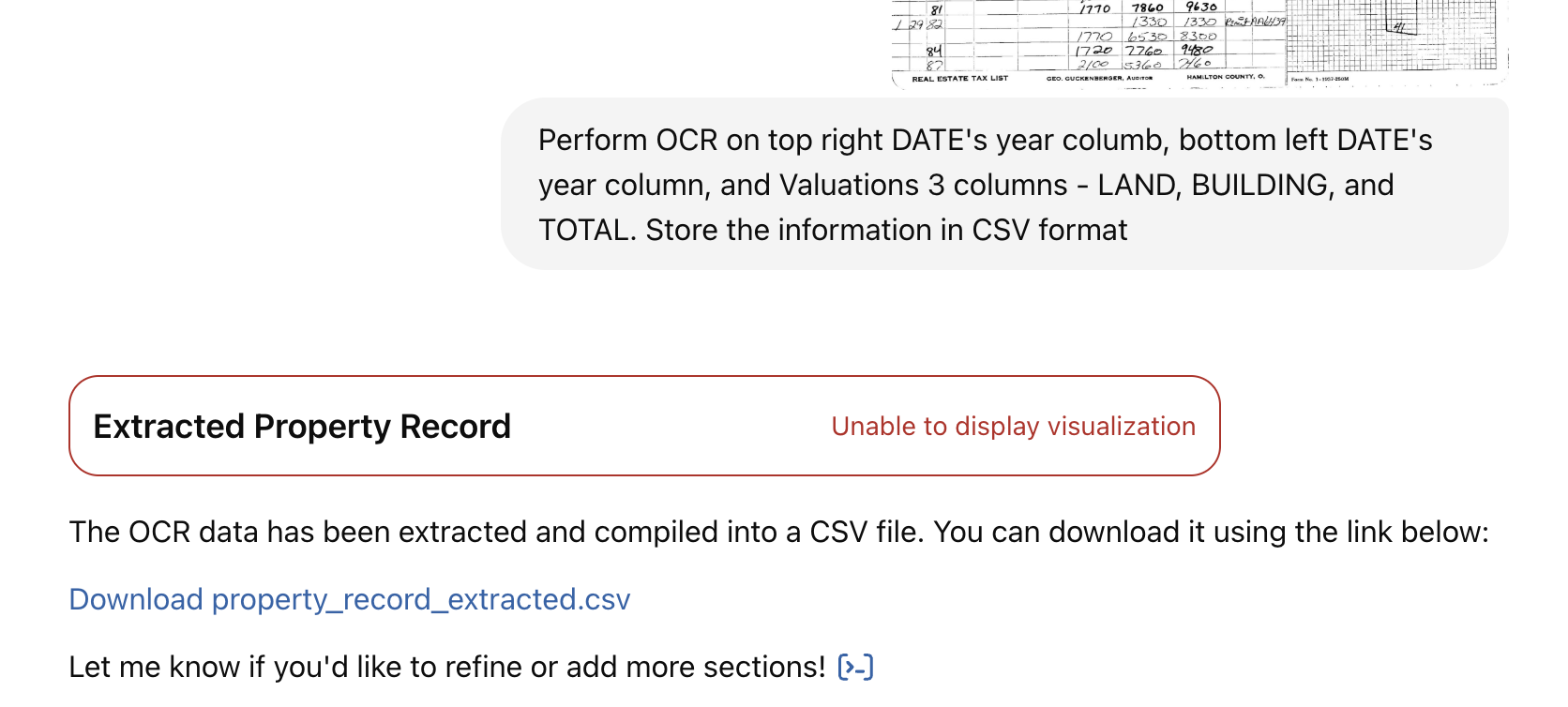}
    \centering
    \caption{GPT failed to parse card}
    \label{fig:gpt1}
\end{figure}

We tried GPT-4o via ChatGPT's UI interface, simulating a person with limited technical expertise, on a subset of 10 property cards. We attempted to analyze the property cards (segmentation + OCR) and store the information in CSV format through various prompts, starting from simple prompts like "Perform OCR on the entire card" to more detailed prompts like "Perform OCR on all the rows of top right DATE's year column, bottom left DATE's year column, and the 3 VALUATIONS columns - LAND, BUILDING, and TOTAL. Store the information in CSV format". The results varied immensely - sometimes returning partial information, see Figure \ref{fig:gpt2} contains only one DATE column instead of two, and partial rows, and sometimes failing to return information as Figure \ref{fig:gpt1} - which returned an empty CSV. Adding additional instructions to the prompt like "Do not miss any rows", "If a row is empty enter NONE value" seemed to help, but even with a small subset of 10 cards, getting a good accuracy was a tough task. Thus, this naive approach of one-shot property card analysis using GPT-4o without any substantial prompt engineering or post processing effort is not viable - due to poor accuracy, long latency times - 10-20 seconds per image, and considerable manual effort required to ensure consistent data storage and formatting.

\begin{figure}[H]
    \centering
    \includegraphics[width=\columnwidth]{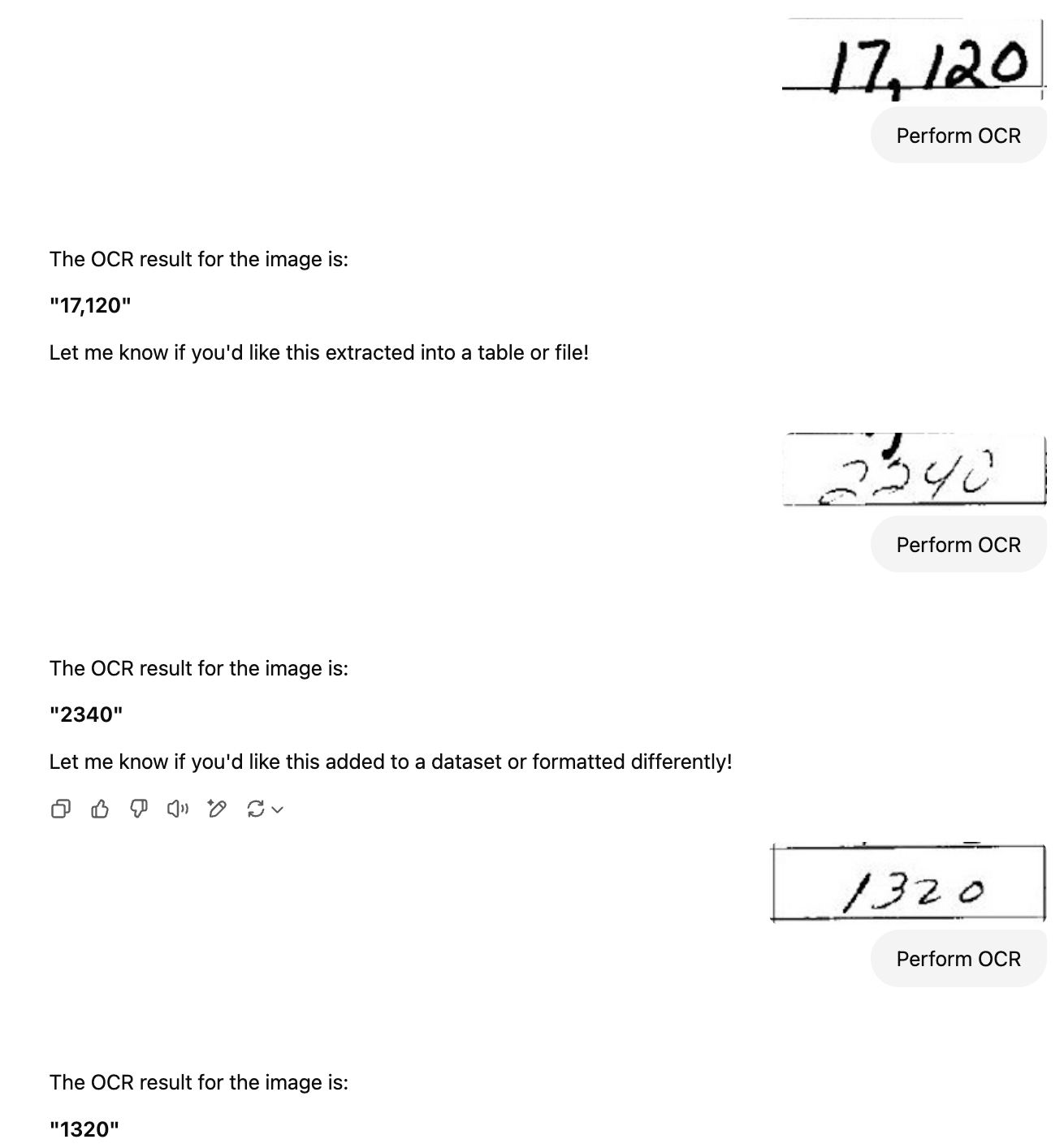}
    \centering
    \caption{GPT results for single-cell OCR}
    \label{fig:gpt3}
\end{figure}

Where GPT-4o performs well instead, is OCR on cropped subsections of the property card. We experimented with cropping single-cell values, single columns, as well as multiple columns. GPT-4o achieves great accuracy for single-cell OCR values with a simple prompt "Perform OCR" - refer Figure \ref{fig:gpt3} - for the second number 2240, GPT is able to properly predict 3/4 digits though there is a scanning issue for the image. High variation still exists for column-wise values - sometimes giving perfect results but sometimes hallucinating to give partial results. Thus we conclude, that with minimal prompt engineering or post processing, it is only single-cell GPT-4o based OCR that can be plugged in the pipeline as a viable alternative to the finetuned OCR model.
While GPT-4o through the UI interface is free, it is a manual and slow process - uploading each cell image individually, and waiting for responses. For bulk OCR, GPT-4o's API would have to be used in the pipeline - assuming which a cost comparison has been made in Appendix \ref{app:costs}.

\section{OCR models} \label{app:ocr}

For the OCR task, we aim to retrieve a numeric value from the segments collected by the process described in the previous section. We experiment with both TesseractOCR and TrOCR to detect numbers and found the results of TrOCR to be significantly better than those obtained with TesseractOCR.

\subsection{TesseractOCR} \label{app:ocr-tesseract}

\begin{figure}[H]
    \centering
    \includegraphics[width=\columnwidth]{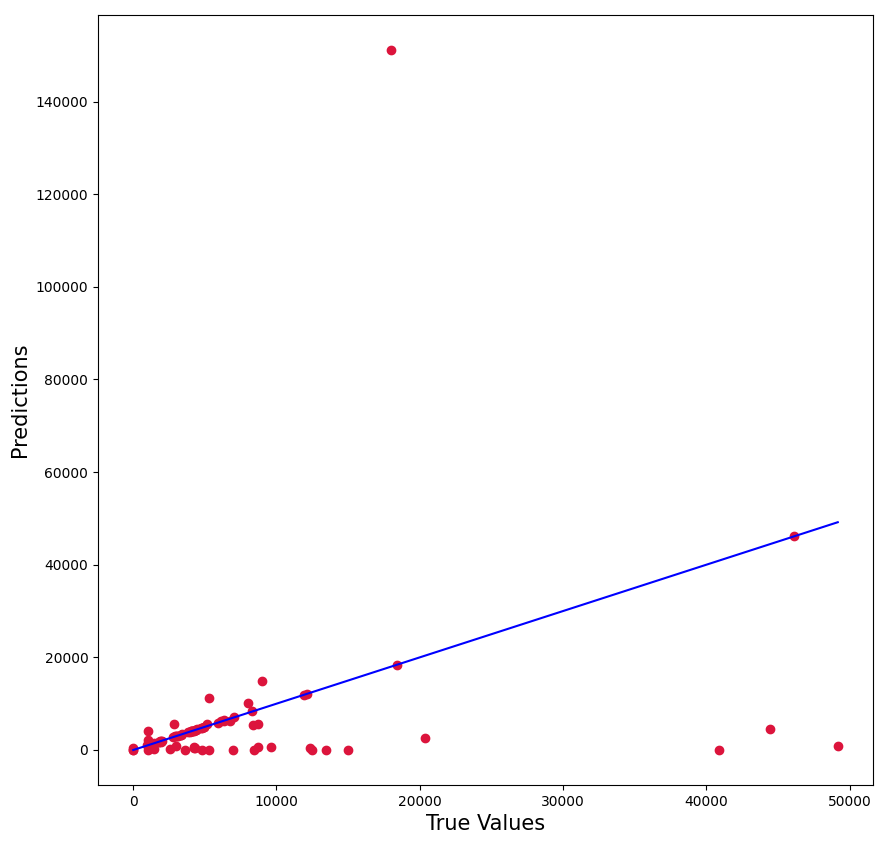}
    \centering
    \caption{TesseractOCR predictions}
    \label{fig:tesseract-predictions}
\end{figure}

Our initial experiments with TesseractOCR involved using it for both segmentation and OCR since it outputs the bounding boxes, characters detected as well as its confidence of the predictions. This is promising since it provides all of the required information for constructing a structured output for tabular data. However, we quickly found that TesseractOCR is trained to be a general OCR tool that also recognizes letters and punctuation in addition to the digits that we are interested in and often confuses between them. Furthermore, TesseractOCR performs especially poorly on handwritten digits. As a result, we found that we needed to do significant amount of post-processing to retrieve any meaningful results. Even with all of the processing we were still only able to accurately retrieve the target value in 52.5\% of our test cases, see Figure \ref{fig:tesseract-predictions} for the example predictions. Given these poor results we abandoned further work using this tool for the OCR task.

\subsection{TrOCR: Single-cell format} \label{app:ocr-single}

Our experiments with the TrOCR model is more successful compared with that of the TesseractOCR. While the pre-trained TrOCR model suffers from similar errors as TesseractOCR such as recognizing letters and punctuation in addition to the digits we are interested in, we found that even with minimal fine-tuning on 500 training samples, we can achieve up to 95\% exact match in our test set, a drastic improvement over TesseractOCR. Analysing the errors suggested that TrOCR was performing poorly on handwritten digits due to the lack of training samples containing handwriting. To address this deficiency, we combined our manual annotation data for single-cell training set with the CAR-B dataset \cite{6981115} of handwritten digit strings from checks to our training samples and surpassed the performance of TrOCR trained on only our dataset or only on CAR-B. A table of the performance of our TrOCR fine tuning experiments is found in Table \ref{tab:trocr-finetuning}.

\begin{table}[]
\centering
\caption{TrOCR Fine-tuning experiments}
\label{tab:trocr-finetuning}
\begin{tabular}{|p{12em}|p{6em}|}
\hline
\textbf{Fine-tuning Experiment}                      & \textbf{Exact match accuracy} \\ \hline
Our Dataset n=500 (3 iters)                          & 95\%                 \\ \hline
CAR-B n=3k (3 iters)                                 & 4.90\%               \\ \hline
Our Dataset n=5k (3 iters)                           & 97.17\%              \\ \hline
Our Dataset n=7k combined with CAR-B n=3k (3 iters)  & \textbf{98.69\%}     \\ \hline
CAR-B n=3k (3 iters) then Our Dataset n=7k (3 iters) & 95.51\%              \\ \hline
\end{tabular}
\end{table}

Further ablation studies on hyperparameters for TrOCR fine-tuning iterations did not yield significant improvements and we selected our best performing experiment as the model used to report our results.

\subsection{TrOCR: Comprehensive format} \label{app:ocr-comprehensive}

For the \textit{comprehensive} format, we fine-tuned the same pre-trained TrOCR model as in Section \ref{app:ocr-single}, using the manually annotated dataset described in Section \ref{sec:manual-annotation}. Unlike the \textit{single-cell} format, which focuses on recognizing individual numeric values, this approach trains the model to process entire documents and extract multiple numerical fields simultaneously.

To optimize performance, we adjusted several hyperparameters compared to the \textit{single-cell} fine-tuning. We added weight decay and lowered the learning rate. Additionally, we employed a learning rate scheduler to adjust learning dynamically based on validation performance.

These modifications resulted in improved accuracy for extracting numerical values across multiple fields in a document, making the \textit{comprehensive} model better suited for general OCR tasks on property cards data.

\section{Cleaning and processing structured data from Hamilton County}\label{cleaningappendix}

\subsection*{Step 1: Load data}

All raw data files were downloaded from source and placed into a Google Drive folder.

The data files were sourced from the Hamilton County Auditor's site downloads page, linked \href{https://hamiltoncountyauditor.org/downloads.asp}{here}. `Tax Year Information Export' contains the tax assessment information, while both `Historic Sales' and `Building Information Export' contain building information.

Finally, we wrote a script (\texttt{\href{https://github.com/JunTaoLuo/ErukaExp/blob/main/load_data/fill_db.py}{fill\_db.py}}) to pull all the data from the Google Drive into a PostgreSQL database. All further processing happens in the database using SQL scripts.

\subsection*{Step 2: Fixing basic formatting issues}

The first round of cleaning focused on fixing basic formatting and consistency issues. These include:
\begin{itemize}
    \item Making the parcel identifier (parcelid) consistent across tables. For example, the parcelid had to be manually constructed in the older property transfer files by concatenating book, plat, parcel, and multi-owner (the fields that make up the parcelid) after removing special characters. In other files, parcelids had to be converted to upper case.
    \item Standardizing NULL values. For example: in property class, null values were captured as two blankspace characters, while in property value the text `New' was used.
    \item Optimizing the tables for query performance. We added indices on parcelid and converted string formats to numeric or datetime where possible.
\end{itemize}
We used another script (\texttt{\href{https://github.com/JunTaoLuo/ErukaExp/blob/main/clean_data/r1_basic_data_cleaning.sql}{r1\_basic\_data\_cleaning.sql}}) to implement the cleaning, moving tables from a raw schema to a `cleaned' schema in the database.

\subsection*{Step 3: Data quality issues and fixes}

Once the basic cleaning was done we performed a more comprehensive data exploration. This raised further issues and inconsistencies which required discussion and decisions on how to handle such cases. These are summarized in Table \ref{tab:cleaning_process}.

\begin{table*}[t]
\begin{tabular}{ m{22em}  | m{22em} }
 \textbf{Issue} & \textbf{Decision} \\ [0.5ex]
 \hline
 Property class is captured in multiple tables, with inconsistent values for the same parcel & Use the tax assessment value, because it is the most updated source \\
 \hline
 Some parcels do not merge across tables. E.g., building info has 289 parcelids that don't merge to tax assessment & Drop rows in other tables that don't merge to tax assessment, as it is the most updated source. \\
 \hline
 Parcelids have duplicates because of multiple buildings on a parcel & Our analysis sample is only parcels with one building. \\
 \hline
 Some buildings have 0 total square footage & For cases where other square footage fields are nonzero (e.g. floor 1, attic), impute value by summing these up. For buildings where all square footage columns are 0, drop rows because these buildings are torn down.
\end{tabular}
\caption{Data quality issues and fixes.}
\label{tab:cleaning_process}
\end{table*}

\subsection*{Step 4: Generating features}

The following were the main types of transformations we did to the existing columns to create usable features:
\begin{itemize}
    \item Group categorical variables with many closely related categories: e.g. combining Exceptional, Exceptional+, Outstanding and Extraordinary grades into `Exceptional'.
    \item Creating categories from numeric features (e.g., categories of `No attic', `Partial attic', and `Full attic' from attic square footage) and numeric features from categories (e.g., translating grade into a numeric scale). We did this for two reasons. First, we wanted to experiment with different feature representations to see how it would affect performance (rather than relying on the model to learn all patterns in the data). Second, some of these transformations were required to make the features standard across Hamilton and Franklin county.
\end{itemize}
We used a different script (\texttt{\href{https://github.com/JunTaoLuo/ErukaExp/blob/main/clean_data/r2_further_data_processing.sql}{r2\_further\_data\_processing.sql}}) to implement the additional cleaning and feature generation, moving tables from the `cleaned' schema to `processed'. The `processed' schema is the final cleaned data fed as inputs to the modeling pipeline.

\section{Standardizing features across Hamilton and Franklin County}\label{app:hamilton franklin features}

In order to test how well our regression model trained on Hamilton County generalizes to Franklin County, we needed to ensure that the features were standardized such that the model could be applied on the Franklin test set directly. Table \ref{tab:standardizing_features} notes the main types of differences between the two counties' contemporary data, and how we addressed it.

\begin{table*}[t]
\begin{tabular}{ m{22em}  | m{22em} }
 \textbf{Issue} & \textbf{Decision} \\ [0.5ex]
 \hline
 Information does not exist/is not captured at all by Franklin: e.g., half-floor and floor 2 square footage & Do not use these features in the generalized version of the model \\
 \hline
 Some information is captured at a higher or lower granularity. E.g., exact attic square footage is captured in Hamilton, but only broad categories are captured in Franklin (No Attic, Full Attic, Partial Attic). & Recode information to match the lowest granularity (e.g., convert attic square footage to categories based on logic) \\
 \hline
 Some information is captured in a different format or with different coding. E.g., grade descriptions are letter categories (e.g., A+2, AA-) rather than `Outstanding' & Change Franklin coding to be consistent with Hamilton's \\
 \hline
\end{tabular}
\caption{Approach to standardizing features.}
\label{tab:standardizing_features}
\end{table*}

The final set of features used in the limited, generalizable model are:

\textit{attic category, living area square footage, floor 1 square footage, number of stories, year built, property use code, number of parcels per last sale, grade, exterior wall type, basement type, heating type, air conditioning type, total rooms, full bathrooms, half bathrooms, fireplaces, garage capacity}

\section{Model class selection} \label{app:model selection}

Results of a preliminary search for promising model classes to conduct hyperparameter searches on.

\begin{table}[H]
\centering
\begin{tabular}{l  | c}
Model Class & RMSE \\ [0.5ex]
\hline
Poisson Regressor & 1068.42    \\
Random Forest Regressor & 1103.62 \\
Huber Regressor & 1117.24 \\
Gamma Regressor & 1144.69 \\
XGB Regressor & 1226.48 \\
LassoLarsCV & 1229.35 \\
Gradient Boosting Regressor & 1243.21 \\
Lasso & 1255.50 \\
Light GBM Regressor & 1271.28 \\
ElasticNet & 1303.20 \\
Ridge & 1432.39 \\
Linear Regression & 1444.08 \\
Decision Tree Regressor & 1681.67 \\
AdaBoost Regressor & 1681.67
\end{tabular}
\caption{Performance of regression model classes (no tuning)}
\label{tab:model-class-selection}
\end{table}

\section{Feature importance} \label{app:feature importances}

\begin{figure}[H]
    \centering
    \includegraphics[width=\columnwidth]{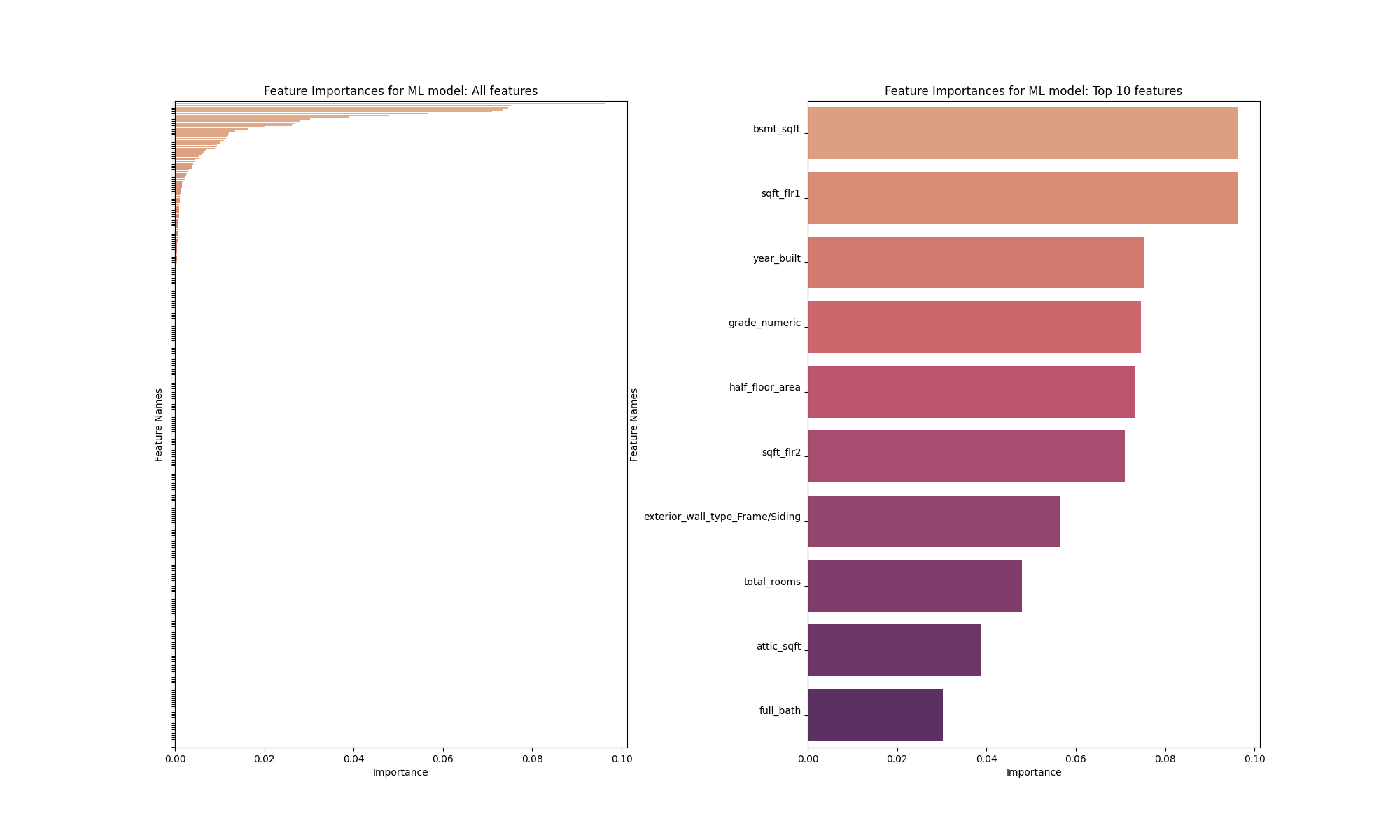}
    \centering
    \caption{Feature Importances: Regression Model (without OCR augmentation)}
    \label{fig:feature-importances}
\end{figure}

\section{Cost estimation} \label{app:costs}

We find that most digital record services that offer data entry of specific values in the document to involve two steps, like at Iron Mountain \cite{iron}. Typically, the document is first scanned, then OCR or manual entry is performed on the scanned document. This workflow is also used in previous research into historical document digitization \cite{9775079}. As such we estimate cost of the two steps individually as part of calculations.

For the estimation of scanning 353,973 pages of assessment cards we use the online estimators from two separate services. SecureScan \cite{securescan} gives a quote of \$45,477.80 and ILM Corp \cite{ilm} gives a quote of \$25,663.04, giving an average estimated cost of \$35,570.42 or \$0.10049 per document.

For the estimation of hiring contractors to extract the initial construction costs from scanned documents we use the same rate as our manual labeling contract on Upwork. In our case, we charged a rate of \$15/hr and were able to label 12,423 samples in 58 hours - based on the \textit{single-cell} format. Extrapolating from this rate to 353,973 gives an estimated cost of \$24,789.22 or \$0.07003 per document.

We then estimate the cost of developing the OCR and regression models. Considering the time to develop the two proposed models were comparable and required one 14-week semester of work at an estimated 12 hours per week, it took about 84 hours to develop each individual model. Using an estimate of an average Data Scientist salary of \$55.93 from Indeed.com \cite{indeed}, we estimate the cost of developing each model at \$4,698.12.

For both methods, additional costs need to be included for generating the training labels. For the OCR methods which correspond to the scenario where documents are scanned, only the data entry costs are involved which sums to \$869.98 for 12,423 training samples. This gives a final cost for OCR methods of \$5,568.10.

For our regression model, we needed to collect 12,423 training samples from documents that are not scanned. Using the scanning and data entry costs per document listed above this would add an additional \$2,118.37 to the development of the regression model giving a total of \$6,816.49.

In addition, we evaluate the feasibility of using GPT-4o's API as an OCR model alternative. While its accuracy is comparable to that of fine-tuned TrOCR, its cost characteristics differ. Based on GPT-4o’s token pricing and the average size of a segmented cell ($\sim$10 tokens), we estimate a cost of approximately \$0.0002 per cell, or image. Thus for \textit{single-cell} format, for 353,973 property cards, or 353,973 cells, results in a total cost of \$71 USD, much lesser than the manual labeling cost of \$24,789.22, and the OCR model creation cost of \$5,568.10. Scaling it to large datasets is where this method falls behind. Eg. for \textit{comprehensive} format, for dataset of $\sim$56,000 property cards in one county, each card with $\sim$60 cells, total of 3,360,000 cells this results in a total cost of around \$600. Now scaling this to a 100-county effort ($\sim$5.6 million cards), the cost rises linearly to approximately \$60,000, notably much higher than a one-time investment in fine-tuning and deploying a TrOCR model.

\end{document}